\pdfoutput=1
\documentclass[11pt,a4paper]{article}
\usepackage{times,latexsym}
\usepackage{url}
\usepackage[T1]{fontenc}
\usepackage[table]{xcolor}
\usepackage{comment}

\usepackage[acceptedWithA]{tacl2021v1}
\usepackage{fontawesome}

\usepackage[utf8]{inputenc}

\usepackage{microtype}

\usepackage[normalem]{ulem}

\usepackage{mathtools}
\usepackage{amssymb}
\usepackage{cleveref}
\usepackage{booktabs}
\usepackage{xspace}
\usepackage{array,multirow,graphicx}
\usepackage{float}
\usepackage{tikz}

\usepackage{pgfplots}

\newcommand{\daffy}[0]{\includegraphics[width=.04\textwidth]{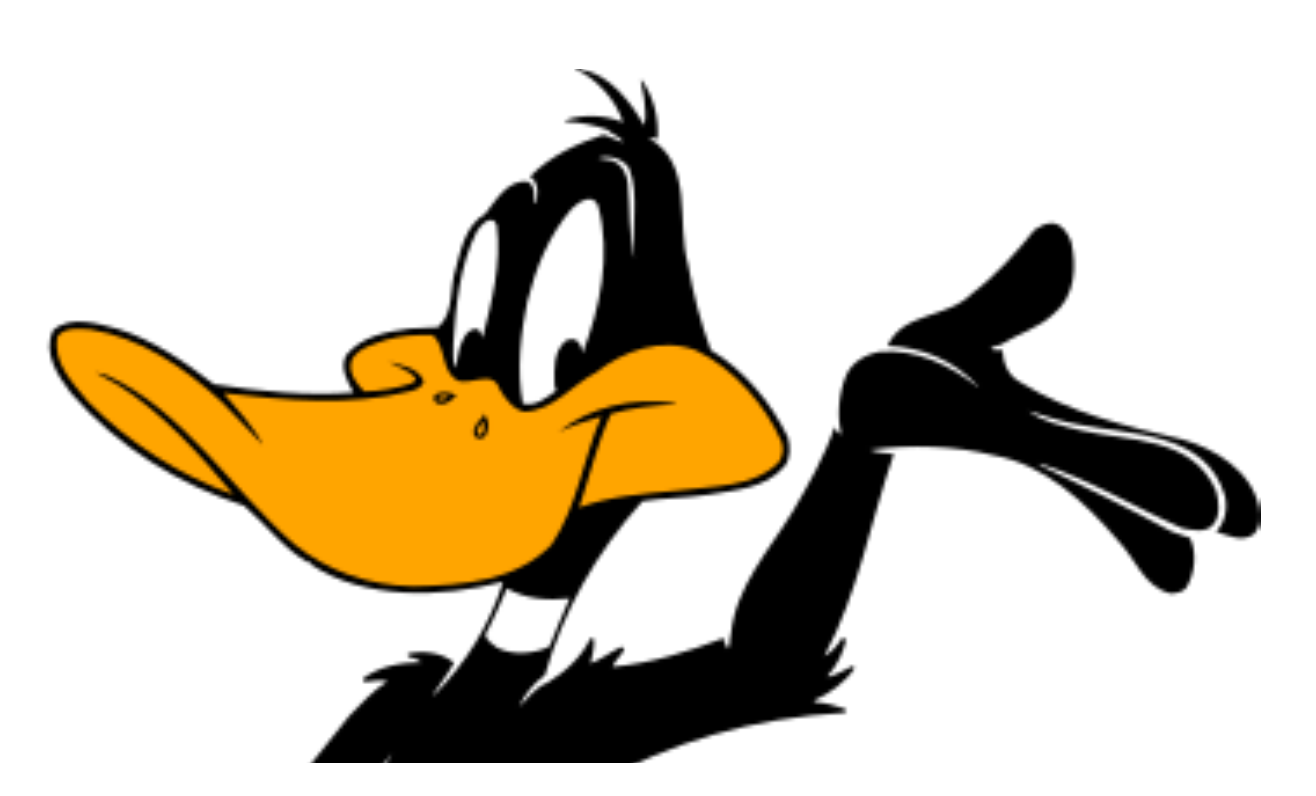}\xspace}

\DeclarePairedDelimiterX\Set[1]\{\}{%

#1
}

\DeclarePairedDelimiterX{\infdivx}[2]{(}{)}{%
  #1\;\delimsize\|\;#2%
}

\newcommand{\CQA}[0]{CQA\xspace}
\newcommand{\DCQA}[0]{CQA\xspace}
\newcommand{\MRQP}[0]{MarCQAp\xspace}

\newcommand{\QuACNH}[0]{QuAC-NH\xspace}

\newcommand{\standard}[0]{\emph{standard}\xspace}
\newcommand{\nohistshort}{\textsc{No Hist.}\xspace}

\newcommand{\nohist}{\textsc{No History}\xspace}
\newcommand{\concat}{\textsc{Concat}\xspace}
\newcommand{\rewrite}{\textsc{Rewrite}\xspace}
\newcommand{\rewritec}{\textsc{Rewrite$_\textit{C}$}\xspace}
\newcommand{\haelf}{HAE$_\textit{LF}$\xspace}
\newcommand{\poshaelf}{PosHAE$_\textit{LF}$\xspace}
\newcommand{\excordlf}{ExCorD$_\textit{LF}$\xspace}
\newcommand{\marqap}[0]{\MRQP}
\newcommand{\marqapc}{\textsc{\mbox{\marqap$_C$}}\xspace}

\newcommand{\concatshort}{\textsc{C}\xspace}
\newcommand{\rewriteshort}{\textsc{R}\xspace}
\newcommand{\rewritecshort}{\textsc{R$_\textit{C}$}\xspace}
\newcommand{\haelfshort}{H\xspace}
\newcommand{\poshaelfshort}{PH\xspace}
\newcommand{\excordlfshort}{Ex\xspace}

\newcommand{\dlt}{$\Delta$\%\xspace}

\newcommand{\rewrittenquestion}{\widetilde{q}_k}
\newcommand{\Qrewritten}{$\rewrittenquestion$\xspace}
\newcommand{\passage}{\widetilde{P}_k}
\newcommand{\Pmarked}{$\passage$\xspace}

\newcommand{\cannot}{\textsc{no answer}}
\newcommand{\hist}{\mathcal{H}}

\newcommand{\Hanswers}[0]{$\{a_i\}_{i=1}^{k-1}$\xspace}

\newcommand{\marqapModel}[0]{\textit{MarCQAp}\xspace}

\newcommand{\random}[0]{\textit{Random Pos}\xspace}
\newcommand{\bare}[0]{\textit{Answer Pos}\xspace}
\newcommand{\semantic}[0]{\textit{Word from $Q$}\xspace}
\newcommand{\semanticfull}[0]{\textit{Full $Q$}\xspace}

\newcommand{\Full}[0]{\textit{Standard}\xspace}
\newcommand{\Ds}[0]{\textit{Domain-Shift}\xspace}
\newcommand{\Lr}[0]{\textit{Low-Resource}\xspace}
\newcommand{\Hr}[0]{\textit{High-Resource}\xspace}

\newcommand{\Nh}[0]{\textit{Noisy-History}\xspace}

\newcommand{\full}[0]{\textit{standard}\xspace}
\newcommand{\ds}[0]{\textit{domain-shift}\xspace}
\newcommand{\lr}[0]{\textit{low-resource}\xspace}
\newcommand{\hr}[0]{\textit{high-resource}\xspace}

\newcommand{\nh}[0]{\textit{noisy-history}\xspace}

\definecolor{cornellred}{rgb}{0.7, 0.11, 0.11}
\newcommand{\red}[1]{{\color{cornellred}#1}}

\definecolor{ballblue}{rgb}{0.13, 0.67, 0.8}
\newcommand{\cand}[1]{{\color{ballblue}#1}}

\definecolor{dartmouthgreen}{rgb}{0.05, 0.5, 0.06}
\newcommand{\green}[1]{{\color{dartmouthgreen}#1}}

\definecolor{amber}{rgb}{1.0, 0.75, 0.0}

\title{On the Robustness of Dialogue History Representation in\\ Conversational Question Answering: A Comprehensive Study \\ and a New Prompt-based Method}

\author{
Zorik Gekhman$^{T}$\Thanks{Authors contributed equally to this work.} \quad
Nadav Oved$^{T}$\footnotemark[\value{footnote}] \quad
Orgad Keller$^{G}$\quad
Idan Szpektor$^{G}$\quad
Roi Reichart$^{T}$\quad \\
$^{T}$Technion - Israel Institute of Technology \quad
$^{G}$Google Research\\
{\tt \{zorik@campus.$|$nadavo@campus.$|$roiri@\}technion.ac.il} \\
{\tt \{orgad$|$szpektor\}@google.com}
}

\date{}

\begin{document}

    \maketitle
    \begin{abstract}

Most works on modeling the conversation history in Conversational Question Answering (\CQA) report a single main result on a common \CQA benchmark. 
While existing models show impressive results on \DCQA leaderboards, it remains unclear whether they are robust to shifts in setting (sometimes to more realistic ones), training data size (e.g. from large to small sets) and domain.
In this work, we design and conduct the first large-scale robustness study of history modeling approaches for \DCQA. 
We find that high benchmark scores do not necessarily translate to strong robustness, and that various methods can perform extremely differently under different settings. Equipped with the insights from our study, we design a novel prompt-based history modeling approach and demonstrate its strong robustness across various settings. 
Our approach is inspired by existing methods that highlight historic answers in the passage. However, instead of highlighting by modifying the passage token embeddings, we add textual prompts directly in the passage text. Our approach is simple, easy-to-plug into practically any model, and highly effective, thus we recommend it as a starting point for future model developers. We also hope that our study and insights will raise awareness to the importance of robustness-focused evaluation, in addition to obtaining high leaderboard scores, leading to better \DCQA systems.\footnote{Our code and data are available at:\\ \url{https://github.com/zorikg/MarCQAp}}

\end{abstract}

    \section{Introduction}
\label{intro}

% \usepackage{graphicx}
% \begin{center}
\begin{table*}[t]
\centering
\small
\scalebox{0.7}{
\begin{tabular}{l|r|r|r|rrrrr|rrr}
\toprule

% Setup
& \multicolumn{1}{c|}{Training} 
& \multicolumn{2}{c|}{In-Domain Evaluation} 
% & \multicolumn{1}{c|}{In-Domain HH} 
% & \multicolumn{1}{c|}{In-Domain MH} 
& \multicolumn{8}{c}{Out-Of-Domain Evaluation}
\\
\midrule

Data source
& \multicolumn{1}{c|}{\multirow{2}{*}{QuAC}} 
& \multicolumn{1}{c|}{\multirow{2}{*}{QuAC}} 
& \multicolumn{1}{c|}{\multirow{2}{*}{QuAC-NH}} 
& \multicolumn{5}{c|}{CoQA}
& \multicolumn{3}{c}{DoQA} 
\\

Domain
&
& 
& 
& \multicolumn{1}{c}{children stories}
& \multicolumn{1}{c}{literature}
& \multicolumn{1}{c}{mid-high school}
& \multicolumn{1}{c}{news}
& \multicolumn{1}{c|}{wikipedia}
& \multicolumn{1}{c}{cooking}
& \multicolumn{1}{c}{movies}
& \multicolumn{1}{c}{travel}
\\
\midrule

\# Examples     
& 83,568
& 7,354
& 10,515
& 1,425
& 1,630
& 1,653
& 1,649
& 1,626
& 1,797
& 1,884
& 1,713
\\

\# Conversations 
&  11,567  
&  1,000  
& 1,204
& 100
& 100
& 100
& 100
& 100
& 400
& 400
& 400 \\
\bottomrule
\end{tabular}
}
\caption{Datasets statistics.}
\label{tab:eval-setups}
% \vspace{-0.5cm}
\end{table*}
% \end{center}

% ----------------------------------------------------------------------

Conversational Question Answering (\DCQA) involves a dialogue between a user who asks questions and an agent that answers them based on a given document.
% \cite{QUAC,COQA,gupta-etal-2020-conversational,DBLP:journals/corr/RogersA21}. 
% \cite{ConvQA_lit_review_2020}.
\DCQA is an extension of the traditional single-turn QA task \cite{SQuAD}, with the major difference being the presence of the conversation history, which requires effective \emph{history modeling} \cite{ConvQA_lit_review_2020}.
Previous work demonstrated that the straightforward approach of concatenating the conversation turns to the input is lacking \cite{HAE}, leading to various proposals of architecture components that explicitly model the conversation history 
\cite{QUAC, FlowQA, FlowDelta, HAE, HAM, GraphFlow, EXCORD}. However, there is no single agreed-upon setting for evaluating the effectiveness of such methods, with the majority of prior work reporting a single main result on a \DCQA benchmark, such as CoQA \cite{COQA} or QuAC \cite{QUAC}.
% The majority of these conversation history modeling methods were evaluated by

While recent \DCQA models show impressive results on these benchmarks, such a single-score evaluation scheme overlooks aspects that can be essential in real-world use-cases. First, QuAC and CoQA contain large annotated training sets, which makes it unclear whether existing methods can remain effective in small-data settings, where the annotation budget is limited. 
In addition, the evaluation is done in-domain, ignoring the model's robustness to domain shifts, with target domains that may even be unknown at model training time. 
Furthermore, the models are trained and evaluated using a ``clean'' 
% and relatively fluent
conversation history between 2 humans, while in reality the history can be ``noisy'' and less fluent, due to the incorrect answers by the model \cite{Ditch_the_Gold_Standard}. Finally, these benchmarks mix the impact of advances in pre-trained language models (LMs) with conversation history modeling effectiveness.

In this work, we investigate the \emph{robustness} of \emph{history modeling} approaches in \DCQA. We ask whether high performance on existing benchmarks also indicates strong robustness. 
To address this question,
we carry out the first large-scale robustness study using 6 common modeling approaches. 
We design 5 robustness-focused evaluation settings, that we curate based on 4 existing \DCQA datasets. Our settings are designed to evaluate efficiency in low-data scenarios, the ability to scale in a high-resource setting, as well as robustness to domain-shift and to noisy conversation history. We then perform a comprehensive robustness study, where we evaluate the considered methods in our settings. 

We focus exclusively on \emph{history modeling}, as it is considered the most significant aspect of \DCQA \cite{ConvQA_lit_review_2020}, differentiating it from the classic single-turn QA task.
To better reflect the contribution of the history modeling component, we adapt the existing evaluation metric.
First, to avoid differences which stem from the use of different pre-trained LMs, we fix the underlying LM for all the evaluated methods, re-implementing all of them.
Second, instead of focusing on final scores on a benchmark, we focus on each model's improvement (\dlt) compared to a baseline QA model that has no access to the conversation history.

Our results show that history modeling methods perform very differently in different settings, and that approaches that achieve high benchmark scores are not necessarily robust under low-data and domain-shift settings. 
% Through careful analysis, we observe that
Furthermore, we notice that
approaches that highlight historic answers within the document by modifying the document embeddings achieve the top benchmark scores, but their performance is surprisingly lacking in low-data and domain-shift settings. 
We hypothesize that history highlighting yields high-quality representation, but since the existing highlighting methods add  dedicated embedding parameters, specifically designed to highlight the document's tokens, they are prone to over-fitting.

These findings motivate us to search for an alternative history modeling approach with improved robustness across different settings. Following latest trends w.r.t.\ prompting in NLP \cite{PROMT_SURVEY}, we design \marqap, a novel prompt-based approach for history modeling, which adds textual prompts within the grounding document in order to highlight previous answers from the conversation history. 
While our approach is inspired by the embedding-based highlighting methods, it is not only simpler, but it also shows superior \emph{robustness} compared to other evaluated approaches.
As \marqap is prompt-based, it can be easily combined with any architecture, 
allowing to fine-tune any model with a QA architecture for the \DCQA task with minimal effort.
Thus, we hope that it will be adopted by the community as a useful starting point, owing to its simplicity, as well as high effectiveness and robustness.
We also hope that our study and insights will encourage more robustness-focused evaluations, in addition to obtaining high leaderboard scores, leading to better \DCQA systems.

    \section{Preliminaries}
\label{sec:preliminaries}

\subsection{\DCQA Task Definition and Notations}
Given a text passage $P$, the current question $q_k$ and a conversation history $\hist_k$ in a form of a sequence of previous questions and answers $\hist_k=(q_1, a_1, \ldots, q_{k-1}, a_{k-1})$, a \DCQA model predicts the answer $a_k$ based on $P$ as a knowledge source. 
The answers can be either spans within the passage $P$ (\emph{extractive}) or free-form text (\emph{abstractive}).

\subsection{\DCQA Datasets}
\label{sec:datasets}

Full datasets statistics are presented in \Cref{tab:eval-setups}.

\paragraph{QuAC} \cite{QUAC} and \textbf{CoQA} \cite{COQA} are the two leading \DCQA datasets, with different properties. In QuAC, the questions are more exploratory and open-ended with longer answers that are more likely to be followed up. This makes QuAC more challenging and realistic.
% \cite{QUAC}. This makes QuAC more challenging.

We follow the common practice in recent works \cite{HAM, HAE, EXCORD, Ditch_the_Gold_Standard}, focusing on QuAC as our main dataset, using its training set for training and its validation set for in-domain evaluation (the test set is hidden, reserved for a leaderboard challenge).
We use CoQA for additional pre-training or for domain-shift evaluation.

\paragraph{DoQA} \cite{DOQA} is another \DCQA dataset with dialogues from the Stack Exchange online forum. 
Due to its relatively small size, it is typically used for testing transfer and zero-shot learning. We use it for domain-shift evaluation.

\paragraph{QuAC Noisy-History (\QuACNH)} 
is based on a datatset of human-machine conversations collected by \citet{Ditch_the_Gold_Standard}, using 100 passages from the QuAC validation set. While \citeauthor{Ditch_the_Gold_Standard} used it for human evaluation, we use it for automatic evaluation, leveraging the fact that the answers are labeled for correctness, which allows us to use the \emph{correct} answers as labels.

In existing \DCQA datasets, each conversation $(q_1, a_1, .. , q_m, a_m)$ and the corresponding passage $P$, are used to create $m$ examples
$\{E_k\}_{k=1}^{m}=\{(P, \hist_k, q_k)\}_{k=1}^{m}$, where $\hist_k=(q_1, a_1, ... q_{k-1}, a_{k-1})$.
$a_k$ is then used as a label for $E_k$. 
Since QuAC-NH contains incorrect answers, if $a_k$ is incorrect we discard $E_k$ to avoid corrupting the evaluation set with incorrectly labeled examples.
We also filtered out invalid questions \cite{Ditch_the_Gold_Standard} and answers that did not appear in $P$.\footnote{Even though \citeauthor{Ditch_the_Gold_Standard} only used extractive models, a small portion of the answers did not appear in the passage.}

\subsection{\DCQA Related Work}
\label{sec:related}

\emph{Conversation History Modeling} is the major challenge in \DCQA \cite{ConvQA_lit_review_2020}. Early work used recurrent neural networks (RNNs) and variants of attention mechanisms \cite{COQA, QUAC, SDNet}. Another trend was to use flow-based approaches, which generate a latent representation for the tokens in $\hist_k$, using tokens from $P$ \cite{FlowQA, FlowDelta, GraphFlow}. Modern approaches, which are the focus of our work, leverage Transformer-based \cite{Transformer} pre-trained language models.

The simplest approach to model the history with pre-trained LMs is to concatenate $\hist_k$ with $q_k$ and $P$ \cite{QUAC, RoR}. 
Alternative approaches rewrite $q_k$ based on $\hist_k$ and use the rewritten questions instead of $\hist_k$ and $q_k$ \cite{Rewrite}, or as an additional training signal \cite{EXCORD}.
Another fundamental approach is to highlight historic answers within $P$ by modifying the passage's token embeddings \cite{HAE, HAM}. 
\citeauthor{HAM} also introduced a component that performs dynamic history selection after each turn is encoded. 
Yet, in our corresponding baseline we utilize only the historic answer highlighting mechanism, owing to its simplicity and high effectiveness. A contemporaneous work proposed a global history attention component, designed to capture long-distance dependencies between conversation turns \cite{GHR}.\footnote{Published 2 weeks before our submission.}

    \section{History Modeling Study}
\label{sec:study}

In this work, we examine the effect of a model's history representation on its robustness.
To this end, we evaluate different approaches under several settings that diverge from the \standard supervised benchmark (\S \ref{sec:setups}). This allows us to examine whether the performance of some methods deteriorates more quickly than others in different scenarios. 
To better isolate the gains from history modeling, we measure performance compared to a baseline QA model which has no access to $\hist_k$ (\S \ref{sec:metric}), and re-implement all the considered methods using the same underlying pre-trained language model (LM) for text representation (\S \ref{pre-trained-lm}).

% ----------------------------------------------------------------------
\begin{table}[t]
\centering
\small
\scalebox{0.7}{
\begin{tabular}
{l|c|c|c}
\toprule

& \multicolumn{1}{c|}{\multirow{1}{*}{Pre-trained}} 
& \multicolumn{1}{c|}{\multirow{2}{*}{Training}} 
& \multicolumn{1}{c}{\multirow{2}{*}{Evaluation}}  
\\ 
& \multicolumn{1}{c|}{\multirow{1}{*}{LM Size}} 
& 
& 
\\ 
\midrule

\Full 
& \multicolumn{1}{c|}{\multirow{1}{*}{Base}}
& \multicolumn{1}{c|}{\multirow{1}{*}{QuAC}} 
& \multicolumn{1}{c}{\multirow{1}{*}{QuAC}} 
\\ 
\midrule

\multicolumn{1}{l|}{\multirow{1}{*}{\Hr}}
& \multicolumn{1}{c|}{\multirow{1}{*}{Large}}
& \multicolumn{1}{c|}{\multirow{1}{*}{CoQA + QuAC}}
& \multicolumn{1}{c}{\multirow{1}{*}{QuAC}} 
\\ 
\midrule

\multicolumn{1}{l|}{\multirow{1}{*}{\Lr}}
& \multicolumn{1}{c|}{\multirow{1}{*}{Base}}
& \multicolumn{1}{c|}{\multirow{1}{*}{QuAC smaller samples}}
& \multicolumn{1}{c}{\multirow{1}{*}{QuAC}} 
\\ 
% & 
% & \multicolumn{1}{c|}{\multirow{1}{*}{samples}}
% &
% \\ 
\midrule

\multicolumn{1}{l|}{\multirow{1}{*}{\Ds}}
& \multicolumn{1}{c|}{\multirow{1}{*}{Base}}
& \multicolumn{1}{c|}{\multirow{1}{*}{QuAC}}
& CoQA + DoQA
\\ 
% &
% &
% & (8 domains) 
% \\ 
\midrule

\Nh 
& \multicolumn{1}{c|}{\multirow{1}{*}{Base}}
& \multicolumn{1}{c|}{\multirow{1}{*}{QuAC}}
& \multicolumn{1}{c}{\multirow{1}{*}{QuAC-NH}}
\\ 
\bottomrule

\end{tabular}
}
\vspace{-0.1cm}
\caption{
Summary of our proposed settings.
}
\label{tab:setups}
\vspace{-0.3cm}
\end{table}

% ----------------------------------------------------------------------

\subsection{Robustness Study Settings}
\label{sec:setups}
We next describe each comparative setting in our study and the rationale behind it, as summarized in \Cref{tab:setups}. \Cref{tab:eval-setups} depicts the utilized datasets.

\paragraph{\Full.}
Defined by \citet{QUAC}, this setting is followed by most works. We use a medium-sized pre-trained LM for each method, commonly known as its \emph{base} version, then fine-tune and evaluate the models on QuAC.

\paragraph{\Hr.} This setting examines the extent to which methods can improve their performance when given more resources. To this end, we use a \emph{large} pre-trained LM, perform additional pre-training on CoQA (with the \DCQA objective), and then fine-tune and evaluate on QuAC.

\paragraph{\emph{\Lr}.}
In this setting, we examine the resource efficiency of the history modeling approaches by reducing the size of the training set. This setting is similar to the \full setting, except that we fine-tune on smaller samples of QuAC's training set. For each evaluated method we train 4 model variants: $20\%$, $10\%$, $5\%$ and $1\%$, reflecting the percentage of training data retained. 
% Some approaches rely on an external model that performs question rewriting (\S \ref{sec:methods}), we choose to treat this model as a given, using the same model for all settings, limiting only the \DCQA training data.
% \footnote{\add{
% % Some approaches are based on a pipeline that first rewrites $q_k$ using an external question rewriting (QR) model (\S \ref{sec:methods}).We use the same QR model across all settings, limiting only the \DCQA data.
% % For question rewriting approaches (see \rewrite and \rewritec in \S \ref{sec:methods}), which rewrite $q_k$ using an external question rewriting (QR) model, we use the same QR model across all settings, limiting only the CQA data.
% Some approaches are based on a pipeline that first rewrites $q_k$ using an external question rewriting (QR) model (see \rewrite and \rewritec in \S \ref{sec:methods}). For these methods we leverage the same QR model across all settings, limiting only the \DCQA data.
% }}

\paragraph{\Ds.} This setting examines \emph{robustness} to domain shift. To this end, we use the $8$ domains in the CoQA and DoQA datasets as test sets from unseen target domains, evaluating the models trained under the \full setting on these test-sets.

\paragraph{\Nh.} This setting examines robustness to noisy conversation history, where the answers are sometimes incorrect and the conversation flow is less fluent. To this end, we evaluate the models trained under the \full setting on the QuAC-NH dataset, consisting of conversations between humans and \emph{other} CQA models (\S \ref{sec:datasets}).
We note that a full human-machine evaluation requires a human in the loop. We choose to evaluate against \emph{other} models predictions as a middle ground. This allows us to test the models' behaviour on noisy conversations with incorrect answers and less fluent flow, but without a human in the loop.

% ----------------------------------------------------------------------

\subsection{Evaluation Metric}
\label{sec:metric}
The standard \DCQA evaluation metric is the average word-level F1 score \cite{SQuAD, QUAC, COQA, DOQA}.\footnote{We follow the calculation presented in \citet{QUAC}.}
Since we focus on the impact of history modeling, we propose to consider each model's improvement in F1 (\dlt) compared to a baseline QA model that has no access to the dialogue history.

% % ----------------------------------------------------------------------
% \begin{table}[t]
% \centering
% \small
% \scalebox{0.64}{
% \begin{tabular}{l|l|l|l}
% \toprule
% Method & Original LM & Original Result & Our Result \\
% \midrule
% \concat \cite{HAE} & BERT & 62.0 & 65.8 \\ 
% \midrule
% \rewrite \cite{Rewrite} & BERT & Not Reported & 64.6 \\ 
% \midrule
% \rewritec (Ours) & \multicolumn{2}{c|}{N/A} & 67.3 \\ 
% \midrule
% ExCorD \cite{EXCORD} & RoBERTa & 67.7 & 67.5 \\ 
% \midrule
% HAE \cite{HAE} & BERT& 63.9 & 68.9 \\ 
% \midrule
% PosHAE \cite{HAM} & BERT & 64.7 & 69.8 \\ 
% % \midrule
% % \excordlf & 56.8 (\green{+13.8\%}) \\ 
% % \haelf & 57.9 (\green{+16.0\%}) \\ 
% % \poshaelf & 60.1 (\green{+20.4\%}) \\ \midrule
% \bottomrule
% \end{tabular}
% }
% \caption{
% \add{Comparison between the original reported F1 scores for all evaluated methods and our re-implementation using the Longformer LM. All results are taken from the \standard setting.}
% }
% \label{tab:sanity-check}
% % \vspace{-0.3cm}
% \end{table}

% % ----------------------------------------------------------------------
% ----------------------------------------------------------------------
% ----------------------------------------------------------------------

% ----------------------------------------------------------------------
\subsection{Pre-trained LM}
\label{pre-trained-lm}
To control for differences which stem from the use of different pre-trained LMs,
% To avoid biases stemming from different pre-trained LMs, 
we re-implement all the considered methods using the Longformer \citep{Longformer}, a sparse-attention Transformer designed to process long input sequences. It is therefore a good fit for handling the conversation history and the source passage as a combined (long) input.
Prior work usually utilized dense-attention Transformers, whose input length limitation forced them to truncate $\hist_k$ and split $P$ into chunks, processing them separately and combining the results \cite{QUAC, HAE, HAM, EXCORD, RoR}. This introduces additional complexity and diversity in the implementation, while with the Longformer we can keep implementation simple, as this model can attend to the entire history and passage.

We would also like to highlight RoR \cite{RoR}, which enhances a dense-attention Transformer to better handle long sequences.
Notably, the state-of-the-art result on QuAC was reported using ELECTRA+RoR with simple history concatenation (see \concat in \S \ref{sec:methods}).
% , in a setting equivalent to our \hr setting (\S \ref{sec:setups}).
While this suggests that ELECTRA+RoR can outperform the Longformer, 
% at least in the \hr setting, 
since our primary focus is on analyzing the robustness of different history modeling techniques rather than on long sequence modeling, we opt for a general-purpose commonly-used LM for long sequences, which exhibits competitive performance.

% ----------------------------------------------------------------------
\subsection{Evaluated Methods}
\label{sec:methods}
In our study we choose to focus on modern history modeling approaches that leverage pre-trained LMs. These models have demonstrated significant progress in recent years (\S \ref{sec:related}).

% We next describe the evaluated methods:

\paragraph{\nohist}
A classic single-turn QA model without access to $\hist_k$.
We trained a Longformer for QA \cite{Longformer}, using $q_k$ and $P$ as a single packed input sequence (ignoring $\hist_k$). 
The model then \emph{extracts} the answer span by predicting its start and end positions within $P$.

In contrast to the rest of the evaluated methods, we do not consider this method as a baseline for history modeling, but rather as a reference for calculating our \dlt metric. As discussed in \S \ref{sec:metric}, we evaluate all history modeling methods for their ability to improve over this model.

\paragraph{\concat}
Concatenating $\hist_k$ to the input (i.e. to $q_k$ and $P$), which is (arguably) the most straightforward way to model the history \cite{QUAC, HAE, RoR}.
Other than the change to the input, this model architecture and training is identical to \nohist. 

\paragraph{\rewrite}
This approach was proposed in \citet{Rewrite}. It consists of a pipeline of two models, question rewriting (QR) and question answering (QA). An external QR model first generates a rewritten question \Qrewritten, based on $q_k$ and $\hist_k$. \Qrewritten and $P$ are then used as input to a standard QA model, identical to \nohist, but trained with the rewritten questions.
For the external QR model we follow \citet{DBLP:journals/corr/LinSC20, Rewrite, EXCORD} and fine-tune T5-base
\cite{T5}
% \footnote{\url{https://huggingface.co/t5-base}}
on the CANARD dataset \cite{CANARD}.
We use the same QR model across all the settings in our study (\S \ref{sec:setups}), meaning that in the \lr setting we limit only the \DCQA data, which is used to train the QA model.

\paragraph{\rewritec}
Hypothesizing that there is useful information in $\hist_k$ on top of the rewritten question \Qrewritten, we combine \rewrite and \concat, obtaining a model which is similar to \concat, except that it replaces $q_k$ with \Qrewritten.

\paragraph{\excordlf}
Our implementation of the ExCorD approach, proposed in \citet{EXCORD}.
Instead of rewriting the original question, $q_k$, at inference time (\rewrite), ExCorD uses the rewritten question only at training time
as a regularization signal when encoding the original question.
% as a signal to guide the model when encoding the original question. 
% While the original work used RoBERTa, we remind the reader that, for stratified comparison, we used the Longformer, observing a comparable result on QuAC (\Cref{tab:main-table}).

\paragraph{\haelf}
Our implementation of the HAE approach proposed in \citet{HAE}, which highlights the conversation history within $P$. Instead of concatenating $\hist_k$ to the input, HAE highlights the historic answers \Hanswers within $P$, by modifying the passage token embeddings. HAE adds an additional dedicated embedding layer with 2 learned embedding vectors, denoting whether a token from $P$ appears in any historic answers or not.

\paragraph{\poshaelf}
Our implementation of the PosHAE approach proposed in \citet{HAM}, which extends HAE by adding positional information. The embedding matrix is extended to contain a vector per conversation turn, each vector representing the turn that the corresponding token appeared in.

% % ----------------------------------------------------------------------
% \begin{table*}[t]
% \centering
% \small
% \scalebox{0.8}{
% \begin{tabular}{l|l|l|l|l}
% \toprule
% & Original Work & Original LM & Original Result & Our Result \\
% \midrule
% \concat & \citet{HAE} & BERT \cite{BERT} & 62.0 & 65.8 \\ 
% \midrule
% \rewrite & \citet{Rewrite} & BERT \cite{BERT} & Not Reported & 64.6 \\ 
% \midrule
% \rewritec & \multicolumn{3}{c|}{N/A (this baseline was first proposed in this work)} & 67.3 \\ 
% \midrule
% ExCorD & \citet{EXCORD} & RoBERTa \cite{ROBERTA} & 67.7 & 67.5 \\ 
% \midrule
% HAE & \citet{HAE} & BERT \cite{BERT} & 63.9 & 68.9 \\ 
% \midrule
% PosHAE & \citet{HAM} & BERT \cite{BERT} & 64.7 & 69.8 \\ 
% % \midrule
% % \excordlf & 56.8 (\green{+13.8\%}) \\ 
% % \haelf & 57.9 (\green{+16.0\%}) \\ 
% % \poshaelf & 60.1 (\green{+20.4\%}) \\ \midrule
% \bottomrule
% \end{tabular}
% }
% \caption{
% \add{Comparison between the original reported F1 scores for all evaluated methods and our re-implementation using the Longformer LM. All results are taken from the \standard setting.}
% }
% \label{tab:sanity-check}
% % \vspace{-0.3cm}
% \end{table*}

% % ----------------------------------------------------------------------

% ----------------------------------------------------------------------
\subsection{Implementation Details}
\label{sec:appendix_implementation}
We fine-tune all models on QuAC for $10$ epochs, employ an accumulated batch size of $640$, a weight decay of $0.01$, and a learning rate of $3\cdot10^{-5}$. 
In the high-resource setup, we also pre-train on CoQA for 5 epochs.
% We limit the input length to $2000$ tokens only during training, padding shorter sequences and truncating longer ones. 
We use a maximum output length of $64$ tokens. 
Following \citet{Longformer}, we set Longformer's global attention to all the tokens of $q_k$.
We use the cross-entropy loss and \emph{AdamW} optimizer \citep{DBLP:journals/corr/KingmaB14,DBLP:conf/iclr/LoshchilovH19}.
Our implementation makes use of the \emph{HuggingFace Transformers} \cite{wolf-etal-2020-transformers},
% \footnote{\url{https://github.com/huggingface/transformers}} 
and \emph{PyTorch-Lightning} libraries.\footnote{\url{https://github.com/PyTorchLightning/pytorch-lightning}}

For the \emph{base} LM 
% (used in \full, \lr, \ds and \nh) 
(used in all settings except \hr) 
we found that a Longformer that was further pre-trained on SQuADv2 \cite{DBLP:conf/acl/RajpurkarJL18},
\footnote{\url{https://huggingface.co/mrm8488/longformer-base-4096-finetuned-squadv2}}
achieved consistently better performance than the base Longformer. Thus, we adopted it as our \emph{base} LM. For the \emph{large} LM (used in the \hr setting) we used Longformer-large.\footnote{\url{https://huggingface.co/allenai/longformer-large-4096}}
% \footnote{\url{https://huggingface.co/allenai/longformer-base-4096}}

% \addnotmand{To validate our results, we perform statistic significance tests. Following \citet{HAM} we use the student’s paired t-test with $p < 0.05$. In each setting, we compare the results of the best performing method to all other methods.}
% \addnotmand{In \S \ref{sec:marqap} we introduce \marqap, a novel history modeling method, and perform statistic significance tests comparing it's results to all other methods in each setting. Following \citet{HAM} we use the student’s paired t-test with $p < 0.05$.}
% \addnotmand{In \S \ref{sec:marqap} we introduce \marqap, a novel history modeling method. We performed statistic significance tests comparing it's results to all other methods in each setting. Following \citet{HAM} we used the student’s paired t-test with $p < 0.05$.}
% \addnotmand{In \S \ref{sec:marqap} we introduce a novel method (\marqap). 
% We performed statistic significance tests \cite{STAT_SIG} comparing it to all other methods in each setting. 
% Following \citet{HAM} we used the student’s paired t-test with $p < 0.05$.}
% Our evaluation also included statistic significance tests \cite{STAT_SIG} comparing it to all other methods in each setting. 
% \addnotmand{In \S \ref{sec:marqap}, we introduce a novel method (\marqap) and perform statistical significance tests \cite{STAT_SIG2}. Following \citet{HAM}, we use the student’s paired t-test with $p < 0.05$, to compare (\marqap) to all other methods in each setting.}
In \S \ref{sec:marqap}, we introduce a novel method (\marqap) and perform statistical significance tests \cite{STAT_SIG, STAT_SIG2}.
Following \citet{HAM}, we use the Student’s paired t-test with $p < 0.05$, to compare \marqap to all other methods in each setting.

% \add{
% In our re-implementation of the evaluated methods, we carefully followed their descriptions and implementation details in the corresponding papers and code bases. A key difference in our implementation is the use of a long sequence Transformer, which removes the need to truncate $\hist_k$ and split $P$ into chunks (\S \ref{pre-trained-lm}). This simplifies the code and avoids diversity in the implementation.}\footnote{\add{The length limit on $\hist_k$ can vary between different works, so is the ranking logic between the different document chunks.}} \add{
% \Cref{tab:sanity-check} compares between our results and the results reported in previous works. In almost all cases we got a higher score (probably since the Longformer performs better than BERT), one exception is ExCorD, where we got a comparable score\footnote{\add{The difference is not statistically significant.}} (probably since the Longformer is actually initialized using RoBERTa's weights \cite{Longformer}).}

% ----------------------------------------------------------------------
\begin{table}[t]
\centering
\small
\scalebox{0.6}{
\begin{tabular}{l|l|l|l|l}
\toprule
& Original Work & Original LM & Original Result & Our Impl. \\
\midrule
\concat & \citet{HAE} & BERT & 62.0 & 65.8 \\ 
\midrule
\rewrite & \citet{Rewrite} & BERT & Not Reported & 64.6 \\ 
\midrule
\rewritec & \multicolumn{3}{c|}{N/A (this baseline was first proposed in this work)} & 67.3 \\ 
\midrule
ExCorD & \citet{EXCORD} & RoBERTa & 67.7 & 67.5 \\ 
\midrule
HAE & \citet{HAE} & BERT& 63.9 & 68.9 \\ 
\midrule
PosHAE & \citet{HAM} & BERT & 64.7 & 69.8 \\ 
% \midrule
% \excordlf & 56.8 (\green{+13.8\%}) \\ 
% \haelf & 57.9 (\green{+16.0\%}) \\ 
% \poshaelf & 60.1 (\green{+20.4\%}) \\ \midrule
\bottomrule
\end{tabular}
}
\vspace{-0.1cm}
\caption{
F1 scores comparison between original implementations and ours (using Longformer as the LM), for all methods described in \S \ref{sec:methods}, in the \standard setting.
}
\label{tab:sanity-check}
\vspace{-0.2cm}
\end{table}

% ----------------------------------------------------------------------

In our re-implementation of the evaluated methods, we carefully followed their descriptions and implementation details as published by the authors in their corresponding papers and codebases. A key difference in our implementation is the use of a long sequence Transformer, which removes the need to truncate $\hist_k$ and split $P$ into chunks (\S \ref{pre-trained-lm}). This simplifies our implementation and avoids differences between methods.\footnote{The maximum length limit of $\hist_k$ varies between different works, as well as how sub-document chunks are handled.} \Cref{tab:sanity-check} compares between our results and those reported in previous works. In almost all cases we achieved a higher score (probably since Longformer outperforms BERT), with the exception of ExCorD, where we achieved a comparable score (probably since Longformer is actually initialized using RoBERTa's weights \cite{Longformer}).

    \section{Results and Analysis}
\label{sec:study-results}

% ----------------------------------------------------------------------
% ----------------------------------------------------------------------
% ----------------------------------------------------------------------
% ----------------------------------------------------------------------
% ----------------------------------------------------------------------
% ----------------------------------------------------------------------
% ----------------------------------------------------------------------
% ----------------------------------------------------------------------

\newcolumntype{?}{!{\vrule width 1.2pt}}

% ----------------------------------------------------------------------

\begin{table*}[t]
\centering
\small
\scalebox{0.77}{
\begin{tabular}{l?lllll?l?l}
\toprule

% \rowcolor[HTML]{B7E1CD} 
% \multicolumn{8}{c}{\emph{In-Domain: Human \& Human Conversations}}  \\
% \midrule

Setting
% & \multicolumn{5}{c?}{Low Resource}
% & \multicolumn{1}{c?}{Full}
% & \multicolumn{1}{c}{High Resource}  \\
& \multicolumn{5}{c?}{\Lr}
& \multicolumn{1}{c?}{\Full}
& \multicolumn{1}{c}{\Hr}  \\
\midrule

\multirow{2}{*}{LM} 
& \multicolumn{5}{c?}{Longformer-base}
& \multicolumn{1}{c?}{Longformer-base}
& \multicolumn{1}{c}{Longformer-large}  \\

& \multicolumn{5}{c?}{Pre-trained SQuAD}
& \multicolumn{1}{c?}{Pre-trained SQuAD}
& \multicolumn{1}{c}{Pre-trained CoQA}  \\
\midrule

Training set size
& $800\ (1\%)$
& $4K\ (5\%)$ 
& $8K\ (10\%)$
& $16K\ (20\%)$ 
& \multicolumn{1}{|c?}{Avg $\Delta$\%}
& $80K\ (100\%)$
& $80K\ (100\%)$
\\ \midrule

\nohist
& 45.0
& 50.0 
& 52.9 
& 55.4 
& \multicolumn{1}{|c?}{--}
& 60.4 
& 65.6 
\\
\midrule

\concat 
& 43.9 (\red{-2.4\%}) 
& 51.2 (\green{+2.4\%}) 
& 53.4 (\green{+0.9\%})
& 57.8 (\green{+4.3\%}) 
& \multicolumn{1}{|c?}{\green{+1.3\%}} 
& 65.8 (\green{+8.9\%})  
& 72.3 (\green{+10.2\%}) \\ 

\rewrite 
& 46.5 (\green{+3.3\%}) 
& 54.0 (\green{+8.0\%}) 
& 56.4 (\green{+6.6\%})
& 59.2 (\green{+6.9\%}) 
& \multicolumn{1}{|c?}{\green{+6.2\%}} 
& 64.6 (\green{+7.0\%})
& 69.0 (\green{+5.2\%}) \\ 

\rewritec 
& 42.3 (\red{-6.0\%}) 
& 54.4 (\green{+8.8\%}) 
& 57.2 (\green{+8.1\%}) 
& 60.6 (\green{+9.4\%}) 
& \multicolumn{1}{|c?}{\green{+5.1\%}} 
& 67.3  (\green{+11.4\%})
& 72.5 (\green{+10.5\%}) \\ 

\excordlf 
& 46.0 (\green{+2.2\%}) 
& 53.0 (\green{+6.0\%}) 
& 57.2 (\green{+8.1\%})  
& 60.3 (\green{+8.8\%})
& \multicolumn{1}{|c?}{\green{+6.3\%}}  
& 67.5 (\green{+11.8\%}) & 
73.8 (\green{+12.3\%}) \\ 

\haelf  
& 44.5 (\red{-1.1\%}) 
& 50.8 (\green{+1.6\%}) 
& 55.0 (\green{+4.0\%}) 
& 59.8 (\green{+7.9\%}) 
& \multicolumn{1}{|c?}{\green{+3.1\%}} 
& 69.0 (\green{+14.2\%}) 
& 73.2 (\green{+11.4\%}) \\ 

% \poshaelf 
% & 40.5 (\red{-10.0\%}) 
% & 51.0 (\green{+2.0\%}) 
% & 55.1 (\green{+4.2\%})
% & 60.9 (\green{+9.9\%})
% & \multicolumn{1}{|c?}{\green{+1.5\%}} 
% & 69.8 (\green{+15.6\%})\starr
% & 74.2 (\green{+12.9\%})\starr \\ 
\poshaelf 
& 40.5 (\red{-10.0\%}) 
& 51.0 (\green{+2.0\%}) 
& 55.1 (\green{+4.2\%})
& 60.9 (\green{+9.9\%})
& \multicolumn{1}{|c?}{\green{+1.5\%}} 
& 69.8 (\green{+15.6\%})
& 74.2 (\green{+12.9\%}) \\ 

\midrule

% \marqapModel 
% & \textbf{48.2} (\green{\textbf{+7.1\%}})
% & \textbf{57.4} (\green{\textbf{+14.8\%}})
% & \textbf{61.0} (\green{\textbf{+15.3\%}}) 
% & \textbf{64.2} (\green{\textbf{+15.9\%}}) 
% & \multicolumn{1}{|c?}{\green{\textbf{+13.3\%}}}
% & \textbf{70.2} (\green{\textbf{+16.2\%}})
% & \textbf{74.7} (\green{\textbf{+13.7\%}}) \\
\marqapModel (\S \ref{sec:marqap})
& \textbf{48.2} (\green{\textbf{+7.1\%}})
& \textbf{57.4} (\green{\textbf{+14.8\%}})
& \textbf{61.3} (\green{\textbf{+15.9\%}}) 
& \textbf{64.6} (\green{\textbf{+16.6\%}}) 
& \multicolumn{1}{|c?}{\green{\textbf{+13.6\%}}}
& \textbf{70.2} (\green{\textbf{+16.2\%}})
& \textbf{74.7} (\green{\textbf{+13.7\%}}) \\

\bottomrule
\end{tabular}
}
\caption{
In-domain F1 and \dlt scores on the full QuAC validation set, for the \full, \hr and \lr settings. 
% (\dlt) results are reported for each score, and color coded for \green{positive} and \red{negative} numbers.
We color coded the \dlt for \green{positive} and \red{negative} numbers.
% \addnotmand{The best method (in bold) performs statistically significantly better than all results expect the ones marked with} \starr \addnotmand{.}
% \add{Best results are highlighted in bold. A single highlighted result is statistically significantly better than all other models. When the difference between the best result and another method is not statistically significant, we highlight the another method as well.}
}
\label{tab:main-table}
\end{table*}
% ----------------------------------------------------------------------

% ----------------------------------------------------------------------

\begin{table*}[t]
\centering
\small
\scalebox{0.65}{
\begin{tabular}{@{}l|lllll|lll|r@{}}

\toprule
% setting & \multicolumn{9}{c}{Domain-Shift (Zero-Shot)} \\
Setting & \multicolumn{9}{c}{\Ds} \\
\midrule
% Dataset
% \multicolumn{1}{l|}{\multirow{2}{*}{Domain}}
\multirow{2}{*}{Domain}
& \multicolumn{5}{c|}{CoQA}
& \multicolumn{3}{c|}{DoQA} 
& \multicolumn{1}{c}{\multirow{2}{*}{Avg $\Delta$\%}} 
\\

& \multicolumn{1}{l}{Children Sto.} & \multicolumn{1}{l}{Literature} & \multicolumn{1}{l}{M/H Sch.} & \multicolumn{1}{l}{News} & \multicolumn{1}{l|}{Wikipedia} & \multicolumn{1}{l}{Cooking} & \multicolumn{1}{l}{Movies} & \multicolumn{1}{l|}{Travel} &
\\
\midrule

\nohistshort & 54.8 & 42.6                           & 50.3                         & 50.1                     & 58.2                          & 46.9                        & 45.0                        & 44.0                         &  \multicolumn{1}{c}{--}    \\
\midrule

\concat     & 62.2 (\green{+13.5\%})              & 48.0 (\green{+12.7\%})                    & 55.3 (\green{+9.9\%})                 & 54.9 (\green{+9.6\%})             & 59.9 (\green{+2.9\%})                  & \textbf{54.8} (\textbf{\green{+16.8\%}})               & \textbf{52.0} (\textbf{\green{+15.6\%}})                & 48.4 (\green{+10\%})                & \green{+11.4\%} \\

\rewrite    & 60.1 (\green{+9.7\%})               & 47.7 (\green{+12.0\%})                    & 55.0 (\green{+9.3\%})                   & 54.8 (\green{+9.4\%})             & 60.9 (\green{+4.6\%})                  & 44.6 (\red{-4.9\%})               & 43.2 (\red{-4.0\%})                & 40.9 (\red{-7.0\%})                & \green{+3.6\%}  \\

\rewritec   & 62.7 (\green{+14.4\%})              & 49.0 (\green{+15.0\%})                      & 56.7 (\green{+12.7\%})                & 55.2 (\green{+10.2\%})            & 59.4 (\green{+2.1\%})                  & 52.0 (\green{+10.9\%})                 & 49.1 (\green{+9.1\%})               & 46.4 (\green{+5.5\%})               & \green{+10.0\%} \\

% \excordlf     & 62.7 (\green{+14.4\%})              & 51.5 (\green{+20.9\%})                  & 58.2 (\green{+15.7\%})                & 57.0 (\green{+13.8\%})              & 63.6 (\green{+9.3\%})                  & 53.7 (\green{+14.5\%})\starr               & 51.1 (\green{+13.6\%})\starr              & 48.6 (\green{+10.5\%})\starr              & \green{+14.1\%} \\
\excordlf     & 62.7 (\green{+14.4\%})              & 51.5 (\green{+20.9\%})                  & 58.2 (\green{+15.7\%})                & 57.0 (\green{+13.8\%})              & 63.6 (\green{+9.3\%})                  & 53.7 (\green{+14.5\%})               & 51.1 (\green{+13.6\%})              & 48.6 (\green{+10.5\%})              & \green{+14.1\%} \\

\haelf     & 61.8 (\green{+12.8\%})             & 50.5 (\green{+18.5\%})             & 56.6 (\green{+12.5\%})             & 55.4 (\green{+10.6\%})             & 60.9 (\green{+4.6\%})             & 45.0 (\red{-4.1\%})             & 45.1 (\green{+0.2\%})             & 45.1 (\green{+2.5\%})              & \green{+7.2\%}  \\

\poshaelf     & 56.6 (\green{+3.3\%})               & 47.4 (\green{+11.3\%})                  & 55.4 (\green{+10.1\%})                & 52.7 (\green{+5.2\%})             & 61.7 (\green{+6.0\%})                    & 45.6 (\red{-2.8\%})               & 45.8 (\green{+1.8\%})               & 44.7 (\green{+1.6\%})               & \green{+4.6\%}  \\
\midrule

% \marqap (\S \ref{sec:marqap})    & \textbf{66.7} (\textbf{\green{+21.7\%}})              & \textbf{56.4} (\textbf{\green{+32.4\%}})                  & \textbf{61.8} (\textbf{\green{+22.9\%}})                & \textbf{60.8} (\textbf{\green{+21.4\%}})            & \textbf{67.5} (\textbf{\green{+16.0\%}})                   & 53.3 (\green{+13.6\%})\starr               & 51.8 (\green{+15.1\%})\starr              & \textbf{50.1} (\textbf{\green{+13.9\%}})              & \textbf{\green{+19.6\%}}
\marqap (\S \ref{sec:marqap})    & \textbf{66.7} (\textbf{\green{+21.7\%}})              & \textbf{56.4} (\textbf{\green{+32.4\%}})                  & \textbf{61.8} (\textbf{\green{+22.9\%}})                & \textbf{60.8} (\textbf{\green{+21.4\%}})            & \textbf{67.5} (\textbf{\green{+16.0\%}})                   & 53.3 (\green{+13.6\%})               & 51.8 (\green{+15.1\%})              & \textbf{50.1} (\textbf{\green{+13.9\%}})              & \textbf{\green{+19.6\%}}

\\
\bottomrule

\end{tabular}
}
\caption{
% Out-of-domain F1 scores for the \ds setting. (\dlt) results are reported for each score, and color coded for \green{positive} and \red{negative} numbers.
F1 and \dlt scores for the \ds setting. We color coded the \dlt for \green{positive} and \red{negative} numbers. 
% \addnotmand{The best method (in bold) performs statistically significantly better than all results expect the ones marked with} \starr \addnotmand{.}
% \add{Best results are highlighted in bold. A single highlighted result is statistically significantly better than all other models. When the difference between the best result and another method is not statistically significant, we highlight the another method as well.}
}
\label{tab:domains}
% \vspace{-0.3cm}
\end{table*}

% ----------------------------------------------------------------------

% ----------------------------------------------------------------------
\begin{table}[t]
\centering
\small
\scalebox{0.65}{
\begin{tabular}{l|l}
\toprule

Setting & \multicolumn{1}{c}{\Nh} \\
\midrule

\nohist & 49.9\\ \midrule
\concat & 55.3 (\green{+10.8\%}) \\ 
\rewrite & 56.0 (\green{+12.2\%}) \\ 
\rewritec & 58.5 (\green{+17.2\%}) \\ 
\excordlf & 56.8 (\green{+13.8\%}) \\ 
\haelf & 57.9 (\green{+16.0\%}) \\ 
\poshaelf & 60.1 (\green{+20.4\%}) \\ \midrule
\marqapModel (\S \ref{sec:marqap}) & \textbf{62.3} (\green{\textbf{+24.9\%}}) \\ 
\bottomrule
\end{tabular}
}
\caption{
F1 and \dlt scores for the \nh setting. 
% \addnotmand{The best method (in bold) performs statistically significantly better than other methods.}
}
\label{tab:noisy-hist}
% \vspace{-0.3cm}
\end{table}

% ----------------------------------------------------------------------
% ----------------------------------------------------------------------
% ----------------------------------------------------------------------
% ----------------------------------------------------------------------

We next discuss the takeaways from our study, where we evaluated the considered methods across the proposed settings.
\Cref{tab:main-table} presents the results of the \full, \hr and \lr settings. \Cref{tab:domains} further presents the \ds results. Finally, \Cref{tab:noisy-hist} depicts the results of the \nh setting.
Each method is compared to \nohist by calculating the \dlt (\S \ref{sec:metric}). The tables also present the results of our method, termed \marqap, which is discussed in \S \ref{sec:marqap}. 

We further analyze the effect of the conversation history length in \Cref{fig:num_turns_plot}, evaluating models from the \full setting with different limits on the history length. 
For instance, when the limit is $2$, we expose the model to up to the $2$ most recent turns, 
% truncating examples with more than $2$ turns to contain only the most recent $2$ turns.
by truncating $\hist_k$.\footnote{We exclude \rewrite, since it utilizes $\hist_k$ only in the form of the rewritten question. For \rewritec, we truncate the concatenated $\hist_k$ for the CQA model, while the QR model remains exposed to the entire history.
}

\paragraph{Key Findings}
A key goal of our study is to examine the robustness of history modeling approaches to setting shifts. This research reveals  limitations of the single-score benchmark-based evaluation adopted in previous works (\S \ref{sec:benchmarks-are-not-enough}), as such scores are shown to be only weakly correlated with low-resource and domain-shift robustness. 
Furthermore, keeping in mind that history modeling is a key aspect of \DCQA, our study also demonstrates the importance of isolating the contribution of the history modeling method from other model components (\S \ref{sec:history-isolation}). Finally, we discover that while existing history highlighting approaches yield high-quality input representations, their robustness is surprisingly poor. We further analyze the history highlighting results and provide possible explanations for this phenomenon (\S \ref{sec:highlighting-is-not-robust}). This finding is the key motivation for our proposed method (\S \ref{sec:marqap}).

% -------------------------------------------------
\subsection{High \DCQA Benchmark Scores do not Indicate Good Robustness}
\label{sec:benchmarks-are-not-enough}
First, we observe some expected general trends: 
All methods improve on top of \nohist, as demonstrated by the \green{positive} \dlt in the \full setting, showing that all the methods can leverage information from $\hist_k$. 
All methods scale with more training data and a larger model (\hr), and their performances drop significantly when the training data size is reduced (\lr) or when they are presented with noisy history. A performance drop is also observed when evaluating on \ds, as expected in the zero shot setting. 

However, not all methods scale equally well and some deteriorate faster than others.
This phenomenon is illustrated in \Cref{tab:methods-ranks}, where the methods are ranked by their scores in each setting. 
We observe high instability between settings. For instance, \poshaelf is top performing in $3$ settings but is second worst in $2$ others. \rewrite is second best in \lr, but among the last ones in other settings. So is the case with \concat: Second best in \ds but among the worst ones in others. In addition, while all the methods improve when they are exposed to longer histories (\Cref{fig:num_turns_plot}), some saturate earlier than others.

We conclude that \textit{the winner does not take it all}: There are significant instabilities in methods' performance across settings. 
This reveals the limitations of the existing single-score benchmark evaluation practice, and calls for more comprehensive robustness-focused evaluation.

% --------------------------------------------------------------------------------
\begin{table}[t]
\resizebox{\columnwidth}{!}{%
\begin{tabular}{ccccc}
\full                              & \hr                     & \lr                      & \ds                      & \nh                     \\

\rowcolor[HTML]{B7E1CD} 
\poshaelfshort \ \green{(+15.6\%)}                        & \poshaelfshort \ \green{(+12.9\%)}                         & \cellcolor[HTML]{F9CB9C}\excordlfshort \ \green{(+6.3\%)} & \cellcolor[HTML]{F9CB9C}\excordlfshort \ \green{(+14.1\%)} & \poshaelfshort \ \green{(+20.4\%)}                         \\

\cellcolor[HTML]{9FC5E8}\haelfshort \ \green{(+14.2\%)}    & \cellcolor[HTML]{F9CB9C}\excordlfshort \ \green{(+12.3\%)} & \cellcolor[HTML]{B4A7D6}\rewriteshort \ \green{(+6.2\%)}  & \cellcolor[HTML]{EA9999}\concatshort \ \green{(+11.4\%)}   & \cellcolor[HTML]{FFE599}\rewritecshort \ \green{(+17.2\%)} \\

\cellcolor[HTML]{F9CB9C}\excordlfshort \ \green{(+11.8\%)}  & \cellcolor[HTML]{9FC5E8}\haelfshort \ \green{(+11.4\%)}     & \cellcolor[HTML]{FFE599}\rewritecshort \ \green{(+5.1\%)}  & \cellcolor[HTML]{FFE599}\rewritecshort \ \green{(+10.0\%)}  & \cellcolor[HTML]{9FC5E8}\haelfshort \ \green{(+16.0\%)}     \\

\cellcolor[HTML]{FFE599}\rewritecshort \ \green{(+11.4\%)}  & \cellcolor[HTML]{FFE599}\rewritecshort \ \green{(+10.5\%)}  & \cellcolor[HTML]{9FC5E8}\haelfshort \ \green{(+3.1\%)}     & \cellcolor[HTML]{9FC5E8}\haelfshort \ \green{(+7.2\%)}     & \cellcolor[HTML]{F9CB9C}\excordlfshort \ \green{(+13.8\%)}  \\

\cellcolor[HTML]{EA9999}\concatshort \ \green{(+8.9\%)}    & \cellcolor[HTML]{EA9999}\concatshort \ \green{(+10.2\%)}    & \cellcolor[HTML]{B7E1CD}\poshaelfshort \ \green{(+1.5\%)}  & \cellcolor[HTML]{B7E1CD}\poshaelfshort \ \green{(+4.6\%)}  & \cellcolor[HTML]{B4A7D6}\rewriteshort \ \green{(+12.2\%)}  \\

\rowcolor[HTML]{B4A7D6} 
\rewriteshort \ \green{(+7.0\%)}                           & \rewriteshort \ \green{(+5.2\%)}                           & \cellcolor[HTML]{EA9999}\concatshort \ \green{(+1.3\%)}    & \rewriteshort \ \green{(+3.6\%)}                           & \cellcolor[HTML]{EA9999}\concatshort \ \green{(+10.8\%)}   
\end{tabular}%
}
\caption{
Per setting rankings of the methods evaluated in our study (top is best), excluding \marqap. \concatshort is \concat, \rewriteshort is \rewrite, \rewritecshort is \rewritec, \excordlfshort is \excordlf, \haelfshort is \haelf and \poshaelfshort is \poshaelf. 
}
\vspace{-0.2cm}
\label{tab:methods-ranks}
\end{table}
% --------------------------------------------------------------------------------

% -------------------------------------------------
\subsection{The Contribution of the History Modeling Method should be Isolated}
\label{sec:history-isolation}

In the \hr setting, \nohist reaches $65.6$ F1, higher than many \DCQA results reported in previous works \cite{QUAC, HAE, HAM, FlowQA}. Since it is clearly ignoring the history, this shows that significant improvements can stem from simply using a better LM. Thus comparing between history modeling methods that use different LMs can be misleading.

This is further illustrated with \haelf's and \poshaelf's results. The score that \citeauthor{EXCORD} reported for ExCorD is higher than \citeauthor{HAE} reported for HAE and PosHAE. While both authors used a setting equivalent to our \full setting, \citeauthor{EXCORD} used RoBERTa while \citeauthor{HAE} used BERT, as their underlying LM. It is therefore unclear whether ExCorD's higher score stems from better history representation or from choosing to use RoBERTa. In our study, \haelf and \poshaelf actually outperform \excordlf in the \full setting. This suggests that these methods can perform better than reported, and demonstrates the importance of controlling for the choice of LM when comparing between history modeling methods.

\begin{figure}
 \centering
\begin{tikzpicture}
\begin{axis}[
    % ticks=none,
    xlabel={\# History Turns},
    ylabel={F1},
    ymin=61, ymax=70.5,
    xmin=1, xmax=11,
    xtick={1,2,3,4,5,6,7,8,9,10,11},
    ytick={61,62,63,64,65,66,67,68,69,70},
    legend pos=north west,
    % ymajorgrids=true,
    grid style=dashed,
    % xmajorgrids,
    % ymajorgrids,
    major grid style={dashed},
    width=7.8cm, height=5.5cm,
    xlabel near ticks, ylabel near ticks,
    x tick label style = {font=\scriptsize},
	y tick label style = {font=\scriptsize},
	ytick style={draw=none},
    xtick style={draw=none},
    % width=4.5cm, height=3.2cm,
%     legend style={at={(0.5,-0.25)},
% 	anchor=north,legend columns=3, font=\scriptsize, draw=none},
	legend style={at={(0.65,0.35)},
	anchor=north,legend columns=2, font=\scriptsize, draw=none, legend cell align={left}},
]
\addplot[
    color=blue,
    % color=\color{cb-blue-green},
    % style={dashed}
    style={line width=1.1pt}
    % style={line width=1.1pt, dashed}
    % mark=*,
    ]
    coordinates {
            (11, 65.7) []
            (10, 65.7) []
            (9, 65.7) []
            (8, 65.7) []
            (7, 65.7) []
    		(6, 65.7) []
    		(5, 65.6) []
			(4, 65.5) []
			(3, 65.2) []
			(2, 64.9) []
			(1, 63.1) []
    };
% \addplot[scatter,only marks, scatter src=explicit symbolic,mark size=4pt, color=blue, mark=*, thick,] coordinates { (6, 65.7) };
\addplot[
    color=cyan,
    % color=\color{Cyan},
    style={line width=1.1pt}
    % style={line width=1.1pt, dashed}
    % style={dashed}
    % mark=*,
    ]
    coordinates {
    		(11, 67.3) []
            (10, 67.3) []
            (9, 67.3) []
            (8, 67.3) []
            (7, 67.3) []
    		(6, 67.3) []
    		(5, 67.3) []
			(4, 67.3) []
			(3, 67.2) []
			(2, 66.9) []
			(1, 65.3) []
    };
\addplot[
    color=brown,
    % color=Bittersweet,
    style={line width=1.1pt}
    % style={line width=1.1pt, dashed}
    % style={dashed}
    % mark=*,
    ]
    coordinates {
    		(11, 67.5) []
            (10, 67.5) []
            (9, 67.5) []
            (8, 67.5) []
            (7, 67.5) []
    		(6, 67.4) []
    		(5, 67.3) []
			(4, 67.1) []
			(3, 66.4) []
			(2, 65.2) []
			(1, 61.9) []
    };
\addplot[
    color=black,
    % color=\color{cb-burgundy},
    style={line width=1.1pt}
    % style={line width=1.1pt, dashed}
    % style={dashed}
    % mark=*,
    ]
    coordinates {
    		(11, 68.9) []
            (10, 68.9) []
            (9, 68.9) []
            (8, 68.9) []
            (7, 68.9) []
    		(6, 68.8) []
    		(5, 68.9) []
			(4, 68.6) []
			(3, 68.3) []
			(2, 67.1) []
			(1, 64.3) []
    };
\addplot[
    color=red,
    style={line width=1.1pt}
    % style={line width=1.1pt, dashed}
    % mark=*,
    ]
    coordinates {
    		(11, 69.8) []
            (10, 69.8) []
            (9, 69.8) []
            (8, 69.7) []
            (7, 69.7) []
    		(6, 69.6) []
    		(5, 69.5) []
			(4, 69.3) []
			(3, 68.9) []
			(2, 68.1) []
			(1, 65.9) []
    };
\addplot[
    color=green,
    style={line width=1.1pt}
    % style={line width=1.1pt, dashed}
    % mark=*,
    ]
    coordinates {
    		(11, 70.2) []
            (10, 70.2) []
            (9, 70.2) []
            (8, 70.1) []
            (7, 70.1) []
    		(6, 70) []
    		(5, 70) []
			(4, 69.8) []
			(3, 69.4) []
			(2, 68.5) []
			(1, 65.5) []
    };
% https://www.iro.umontreal.ca/~simardr/pgfplots.pdf
% \addplot+[only marks, scatter src=explicit symbolic,mark size=2pt, color=blue, color=blue, fill=blue, mark=*, thick] coordinates {
%         (6 ,65.7)};
\addplot+[
        only marks, 
        scatter src=explicit symbolic, 
        mark size=2pt, 
        color=blue, 
        mark=*,
        mark options={solid,fill=blue},
        thick] coordinates {
        (6 ,65.7)};
\addplot+[only marks, scatter src=explicit symbolic,mark size=2pt, 
color=cyan, mark=*, mark options={solid,fill=cyan},
thick] coordinates {
        (4 ,67.3)};
\addplot+[only marks, scatter src=explicit symbolic,mark size=2pt, 
color=brown, mark=*, mark options={solid,fill=brown},
thick] coordinates {
        (7 ,67.5)};
\addplot+[only marks, scatter src=explicit symbolic,mark size=2pt, 
color=black, mark=*, mark options={solid,fill=black},
thick] coordinates {
        (5 ,68.9)};
\addplot+[only marks, scatter src=explicit symbolic,mark size=2pt, 
color=red, mark=*, mark options={solid,fill=red},
thick] coordinates {
        (9 ,69.8)};
\addplot+[only marks, point meta=explicit symbolic, mark size=2pt,
color=green, mark=*, mark options={solid,fill=green},
thick] coordinates {
        (9 ,70.2)};
\legend{\concat, \rewritec, \excordlf, \haelf, \poshaelf, \marqap}
\end{axis}
\end{tikzpicture}
\caption{
    F1 as a function of \# history turns, for models from the \full setup. The first occurrence of the maximum F1 value (saturation point) is highlighted.
    }
    \label{fig:num_turns_plot}
% \vspace{-2cm}
\end{figure}
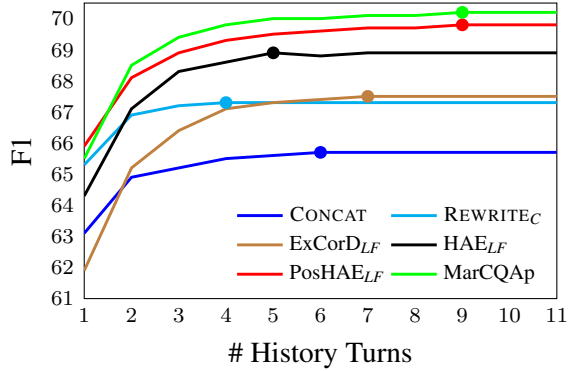

As can be seen in \Cref{fig:num_turns_plot}, \concat saturates at $6$ turns, which is interesting since \citet{HAE} reported saturation at $1$ turn in a BERT-based equivalent. Furthermore, \citeauthor{HAE} observed a performance degradation with more turns, while we observe stability. These differences probably stem from the history truncation in BERT, due to the input length limitation of dense attention Transformers. This demonstrates the advantages of sparse attention Transformers for history modeling evaluation, since the comparison against \concat can be more ``fair''. This comparison is important, since the usefulness of any method should be established by comparing it to the straight-forward solution, which is \concat in case of history modeling. 

% % Begin turn num plot --------------------------------------------------------------------------------
% % End turn num plot --------------------------------------------------------------------------------

We would also like to highlight \poshaelf's F1 scores in the \nh ($60.1$) and the $20\%$ \lr setting ($60.9$), both lower than the $69.8$ F1 in the \full setting. 
\textit{Do these performance drops reflect lower effectiveness in modeling the conversation history?} Here the \dlt comes to the rescue. 
While the \dlt decreased between the \full and the $20\%$ settings (\green{$15.6$} $\rightarrow$ \green{$9.9$}), it actually increased in the \nh setting (to \green{$20.4$}). This indicates that even though the F1 decreased, the ability to leverage the history 
actually increased.

We conclude that our study results support the design choices we made, in our effort to better isolate the contribution of the history representation. We recommend future works to compare history modeling methods using the same LM (preferably a long sequence LM), and to measure a \dlt compared to a \nohist baseline.

% -------------------------------------------------
\subsection{History Highlighting is Effective in Resource-rich Setups, but is not Robust}
\label{sec:highlighting-is-not-robust}

The most interesting results are observed for the history highlighting methods: HAE and PosHAE. 

First, when implemented using the Longformer, \haelf and \poshaelf perform better than reported in previous work, with $68.9$ and $69.8$ F1 respectively, compared to $63.9$ and $64.7$ reported by \citeauthor{HAM} using BERT. The gap between \haelf and \poshaelf demonstrates the effect of the positional information in \poshaelf. This effect is further observed in \Cref{fig:num_turns_plot}, \haelf saturates earlier since it cannot distinguish between different conversation turns, which probably yields conflicting information. \poshaelf saturates at $9$ turns, later than the rest of the methods, which indicates that it can better leverage long conversations.

% Begin data size plot --------------------------------------------------------------------------------
\begin{figure}
 \centering
\begin{tikzpicture}
\begin{axis}[
    xlabel={\# Training Examples},
    ylabel={$\Delta$\%},
    ymin=-15, ymax=18,
    xmin=1, xmax=5,
    xtick={1,2,3,4,5},
    xticklabels={80K, 16K, 8K, 4K, 800},
    ytick={-10,-5,0,5,10,15},
    legend pos=north west,
    % ymajorgrids=true,
    grid style=dashed,
    % xmajorgrids,
    % ymajorgrids,
    major grid style={dashed},
    width=7.8cm, height=5.5cm,
    xlabel near ticks, ylabel near ticks,
    x tick label style = {font=\scriptsize},
	y tick label style = {font=\scriptsize},
	ytick style={draw=none},
    xtick style={draw=none},
% 	axis line style={-},          %% <--- here
%     ylabel style={rotate=-90},    %%<--- here
    % width=4.5cm, height=3.2cm,
    legend style={at={(0.36,0.45)},
	anchor=north,legend columns=2, font=\scriptsize, draw=none, legend cell align={left}},
	y label style={at={(axis description cs:-0.07,.5)},anchor=south},
]

\addplot[
    color=blue,
    % color=\textcolor{cb-blue},
    % style={dashed, line width=1pt}
    style={line width=0.5pt}
    % style={}
    % mark=square,
    ]
    coordinates {
    		(1, 8.9) []
			(2, 4.3) []
			(3, 0.9) []
			(4, 2.4) []
			(5, -2.4) []
    };
    \addlegendentry{\concat}
\addplot[
    color=orange,
    % style={dashed, line width=1pt}
    style={line width=0.5pt}
    ]
    coordinates {
    		(1, 7) []
			(2, 6.9) []
			(3, 6.6) []
			(4, 8) []
			(5, 3) []
    };
    \addlegendentry{\rewrite}    
\addplot[
    color=cyan,
    % style={dashed, line width=1pt}
    style={line width=0.5pt}
    ]
    coordinates {
    		(1, 11.4) []
			(2, 9.4) []
			(3, 8.1) []
			(4, 8.8) []
			(5, -6) []
    };
    \addlegendentry{\rewritec}
\addplot[
    color=brown,
    % style={dashed, line width=1pt}
    style={line width=0.5pt}
    ]
    coordinates {
    		(1, 11.8) []
			(2, 8.8) []
			(3, 8.1) []
			(4, 6.0) []
			(5, 2.2) []
    };
    \addlegendentry{\excordlf}
\addplot[
    color=black,
    % style={dashed, line width=1pt}
    style={line width=0.5pt}
    ]
    coordinates {
    		(1, 14.2) []
			(2, 7.9) []
			(3, 4) []
			(4, 1.6) []
			(5, -1.1) []
    };
    \addlegendentry{\haelf}        
\addplot[
    color=red,
    style={line width=2pt}
    ]
    coordinates {
    		(1, 15.6) []
			(2, 9.9) []
			(3, 4.2) []
			(4, 2) []
			(5, -10) []
    };
    \addlegendentry{\poshaelf}
\addplot[
    color=green,
    style={line width=2pt}
    ]
    coordinates {
    		(1, 16.2) []
			(2, 16.6) []
			(3, 15.9) []
			(4, 14.8) []
			(5, 7.1) []
    };
    \addlegendentry{\marqap}                
\end{axis}
\end{tikzpicture}
\caption{
    $\Delta$\% as a function of \# training examples. Results taken from the \full and \lr settings.
    }
\label{fig:num_examples_plot}
% \vspace{-0.5cm}
\end{figure}
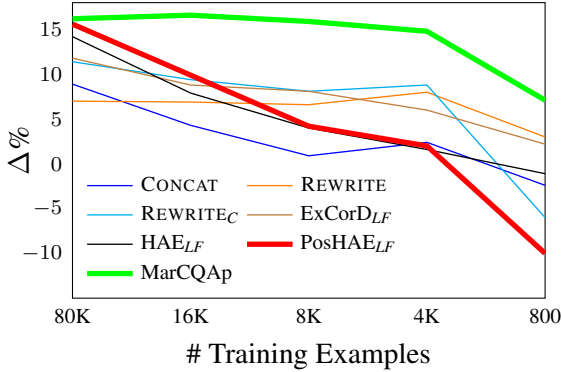
% % End data size plot --------------------------------------------------------------------------------

\poshaelf outperforms all methods in the \full, \hr and \nh settings,\footnote{We ignore \marqap's results in this section.} demonstrating the high effectiveness of history highlighting. However, it shows surprisingly poor performance in \lr and \ds settings, with extremely low average \dlt compared to other methods. The impact of the training set size is further illustrated in \Cref{fig:num_examples_plot}. We plot the \dlt as a function of the training set size, and specifically highlight \poshaelf in bold red. Its performance deteriorates significantly faster than others when the training set size is reduced. In the $1\%$ setting it is actually the worst performing method.

This poor robustness could be caused by the additional parameters added in the embedding layer of \poshaelf. \Cref{fig:num_examples_plot} demonstrates that properly training these parameters, in order to benefit from this method's full potential, seems to require large amounts of data. Furthermore, the poor \ds performance indicates that, even with enough training data, this embedding layer seems to be prone to overfitting to the source domain. 

We conclude that history highlighting clearly yields a very strong representation, but the additional parameters of the embedding layer seem to require large amounts of data to train properly and over-fit to the source domain. \textit{Is there a way to highlight historic answers in the passage, without adding dedicated embedding layers?} 

In \S \ref{sec:marqap} we present \marqap, a novel history modeling approach that is inspired by PosHAE, adopting the idea of history highlighting. However, instead of modifying the passage embedding, we highlight historic answers by adding textual prompts directly in the input text. By leveraging prompts, we reduce model complexity and remove the need for training dedicated parameters, hoping to mitigate the robustness weaknesses of PosHAE.

    \section{\marqap}
\label{sec:marqap}

Motivated by our findings, we design \marqap, a novel prompt-based history modeling approach that highlights answers from previous conversation turns by inserting textual prompts in their respective positions within $P$. By highlighting with prompts instead of embedding vectors, we hope to encode valuable dialogue information, while reducing the learning complexity incurred by the existing embedding-based methods.
Thus, we expect \marqap to perform well not only in high-resource settings, but also in low-resource and domain adaptation settings, in which prompting methods have shown to be particularly useful \cite{GPT3, le-scao-rush-2021-many, PADA}. 

\begin{figure}[t]
 \centering
\includegraphics[width=\columnwidth]{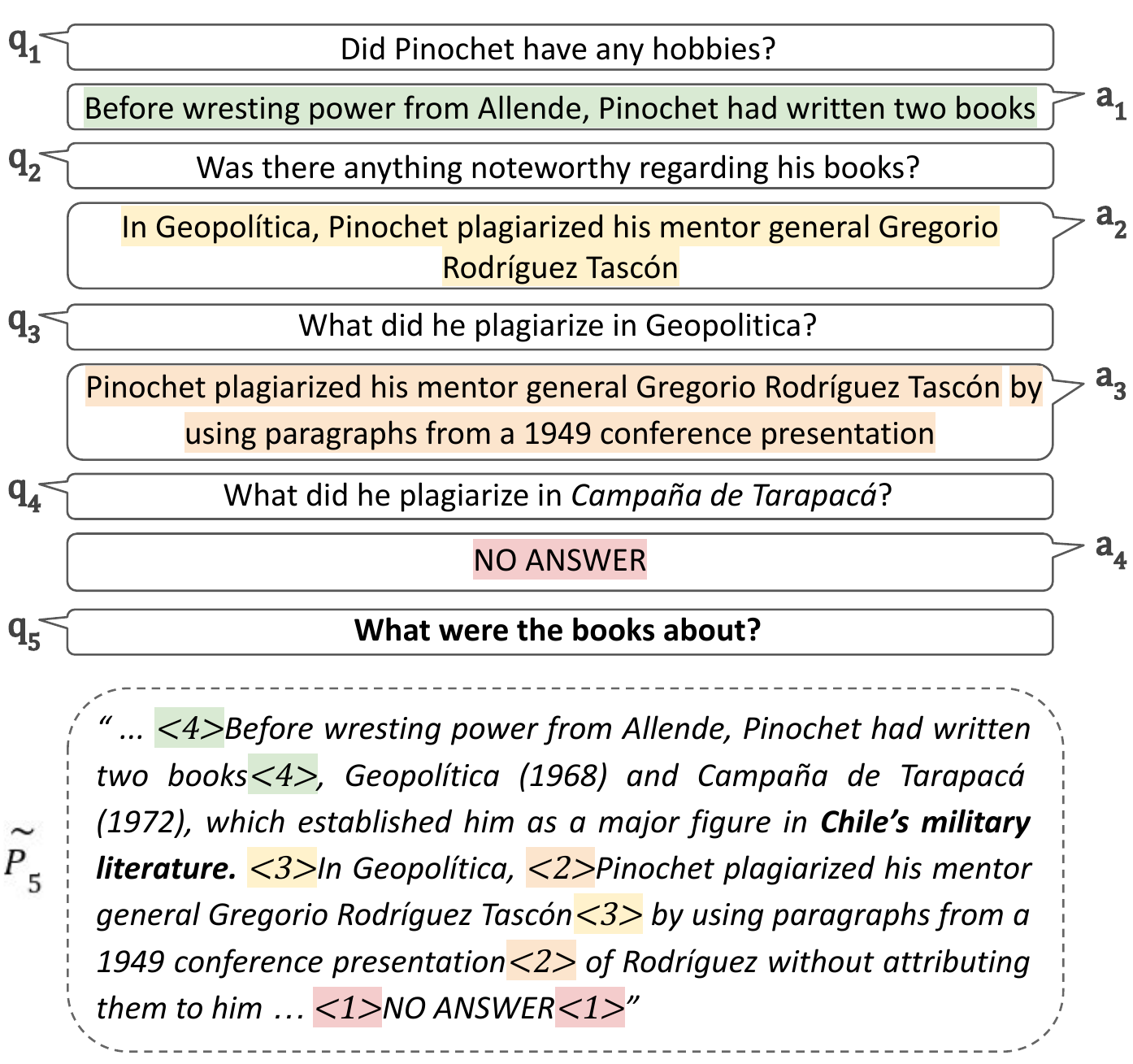}   
    \vspace{-0.7cm}
    \caption{The \MRQP highlighting scheme: Answers to previous questions are highlighted in the grounding document, which is then provided as input to the model.}
    \label{fig:method}    
\vspace{-0.2cm}
\end{figure}

\emph{Prompting} often refers to the practice of adding phrases to the input, in order to encourage pre-trained LMs to perform specific tasks \cite{PROMT_SURVEY}, yet it is also used as a method for injecting task-specific guidance during fine-tuning \cite{le-scao-rush-2021-many, PADA}.
\marqap closely resembles the prompting approach from \citet{PADA} 
since our prompts are: (1) discrete (i.e the prompt is an actual text-string), (2) dynamic (i.e example-based), and (3) added to the \emph{input} text and the model then makes predictions conditioned on the modified input. Moreover, as in \citeauthor{PADA}, in our method the underlying LM is further trained on the downstream task with prompts.
However, in contrast to most prompting approaches, which predefine the prompt's location in the input \cite{PROMT_SURVEY}, our prompts are inserted in different locations for each example.
% These locations highlight previous answers of the CQA model within the document.
In addition, while most \emph{textual} prompting approaches leverage prompts comprised of natural language, our prompts contain non-verbal symbols (e.g\ "\emph{<1>}", see \Cref{fig:method} and \S \ref{sec:method}), which were proven useful for supervision of NLP tasks. 
For instance, \citet{HTLM} showed the usefulness of structured pre-training by adding HTML symbols to the input text.
Finally, to the best of our knowledge, this work is the first to propose a prompting mechanism for the \DCQA task.

% --------------------------------------------------------------------------------
% --------------------------------------------------------------------------------

% ------------------------------------------------------
\subsection{Method}
\label{sec:method}
\marqap utilizes a standard \emph{single-turn} QA model architecture and input, with the input comprised of the current question $q_k$ and the passage $P$. For each \DCQA example $(P, \hist_k, q_k)$, \marqap inserts a textual prompt within $P$, based on information extracted from the conversation history $\hist_k$.
In extractive QA, the answer $a_k$ is typically a span within $P$. 
Given the input $(P, \hist_k, q_k)$, \marqap transforms $P$ into an answer-highlighted passage \Pmarked , by constructing a prompt $p_k$ and inserting it within $P$.
$p_k$ is constructed by locating the beginning and end positions of all historic answers \Hanswers within $P$, and inserting a unique textual marker for each answer in its respective positions (see example in \Cref{fig:method}). 
The input $($\Pmarked$, q_k)$ is then passed to the QA model, instead of $(P, q_k)$.

In abstractive QA, a free form answer is generated based on an evidence span that is first extracted from $P$. Hence, the final answer does not necessarily appear in $P$. To support this setting, \marqap highlights the historical evidence spans (which appear in $P$) instead of the generated answers.

To encode positional dialogue information, the markers for $a_j$ $\in$ \Hanswers include its turn index number in reverse order, i.e. $k-1-j$. 
This encodes relative historic positioning w.r.t.\ the current question $q_k$, allowing the model to distinguish between the historic answers by their recency.

\marqap highlights only the historic answers, since the corresponding questions do not appear in $P$.
While this might lead to information loss, in \S \ref{sec:ablations} we implement \marqap's variants that add the historic questions to the input, and show that the contribution of the historic questions to the performance 
is minor.\footnote{Which is also in line with the findings in \citet{HAM}.}

A \DCQA dialogue may also contain unanswerable questions. Before inserting the prompts, \marqap first appends a `\cannot' string to $P$.\footnote{Only if it is not already appended to $P$, in some datasets the passages are always suffixed with `\cannot'.} Each historical `\cannot' is then highlighted with prompts, similarly to ordinary historical answers. For example see $a_4$ in \Cref{fig:method}.

\marqap has several advantages over prior approaches. 
First, since it is prompt-based, it does not modify the model architecture, which makes it easier to port across various models, alleviating the need for model-specific implementation and training procedures.
Additionally, it naturally represents overlapping answers in $P$, which was a limitation in prior work \cite{HAE,HAM}.
Overlapping answers contain tokens which relate to multiple turns, yet the existing \emph{token-based} embedding methods encode the relation of a token from $P$ only to a single turn from $\hist_k$. Since \marqap is \emph{span-based}, it naturally represents overlapping historic answers (e.g.\ see $a_2$ and $a_3$ in \Cref{fig:method}). 
\begin{comment}
\cand{Finally, \marqap simplifies the input structure and length, since it alleviates the need for including $\hist_k$.}
\end{comment}

% ------------------------------------------------------
\subsection{\marqap Evaluation}
\label{sec:marcqap-eval}
We evaluate \marqap in all our proposed experimental settings (\S \ref{sec:setups}).
As presented in tables \ref{tab:main-table}, \ref{tab:domains} and \ref{tab:noisy-hist}, it outperforms all other methods in all  settings. 
In the \full, \hr and \nh settings, its performance is very close to \poshaelf,
\footnote{In the \standard and \hr \marqap's improvements over \poshaelf are not statistically significant.}
indicating that our prompt-based approach is an effective alternative implementation for the idea of highlighting historical answers. 
Similarly to \poshaelf, \marqap is able to handle long conversations and its performance gains saturate at 9 turns (\Cref{fig:num_turns_plot}). 
However, in contrast to \poshaelf, \marqap performs especially well in the \lr and the \ds settings.

% --------------------------------------------------------------------------------
\begin{figure}[t]
 \centering
 \includegraphics[width=\columnwidth]{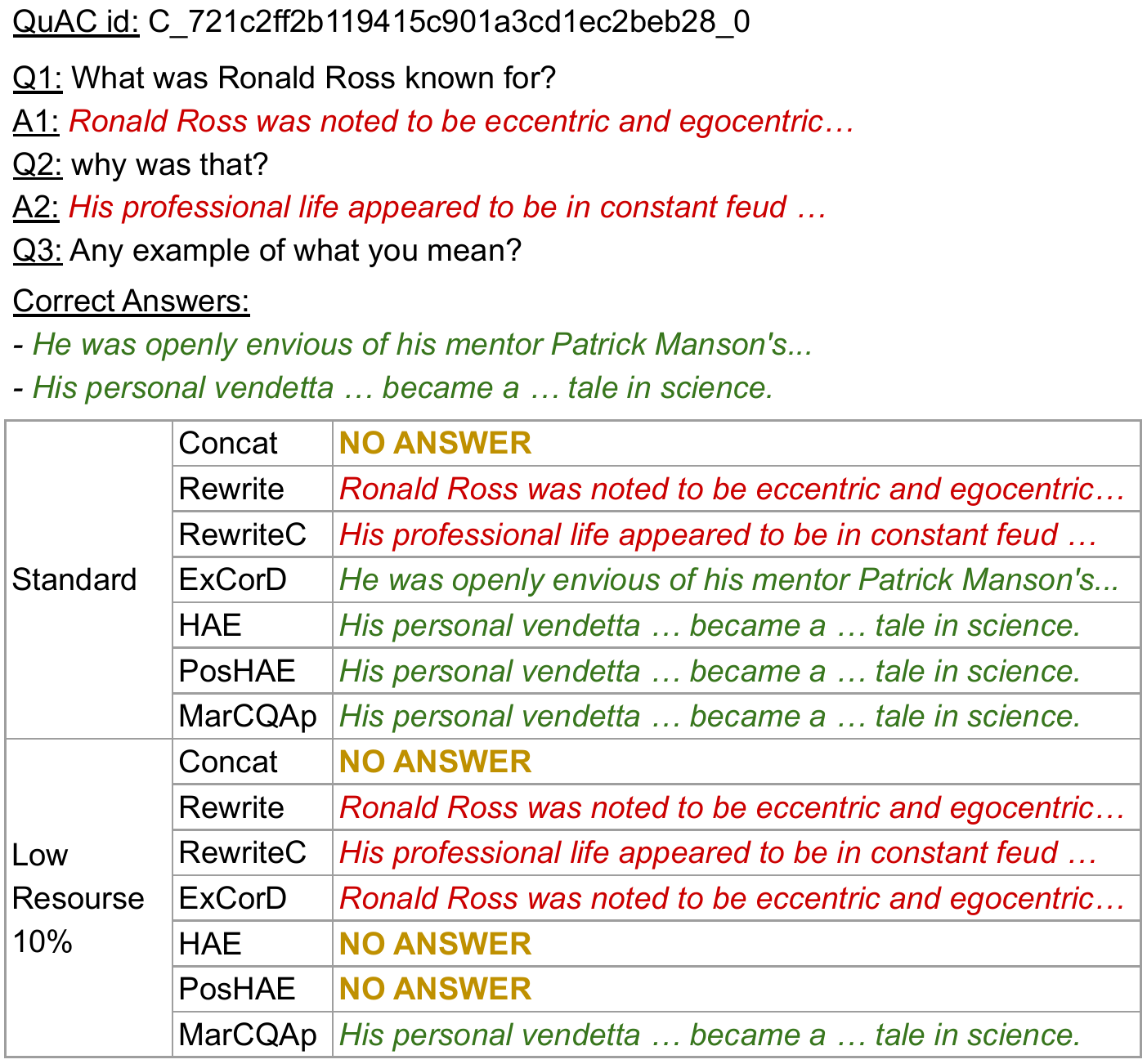}   
    \vspace{-0.7cm}
    \caption{An example of \marqap's robustness in the \lr setting. Even though \excordlf, \haelf and \poshaelf predict correct answers in the \standard setting, they fail on the same example when the training data size is reduced to $10\%$. \marqap predicts a correct answer in both settings.}
    \label{fig:ex2a}    
\vspace{-0.3cm}
\end{figure}
% --------------------------------------------------------------------------------

In the \lr settings, \marqap outperforms all methods by a large margin, with an average \dlt of \green{$13.6\%$} compared to the best baseline with \green{$6.3\%$}.
The dramatic improvement over \poshaelf's average \dlt (\green{$1.5\%$} $\rightarrow$ \green{$13.6\%$}), serves as a strong indication that our prompt-based approach is much more robust. This boost in robustness is best illustrated in \Cref{fig:num_examples_plot}, which presents the \dlt as a function of the training set size, highlighting \poshaelf (red) and \marqap (green) specifically. An example of \marqap's robustness in the \lr setting is provided in \Cref{fig:ex2a}.

In the \ds settings, \marqap is the best performing method in 6 out of 8 domains.\footnote{For the Travel domain \marqap's improvement over \excordlf is not statistically significant.} On the remaining two domains (Cooking \& Movies), \concat is the best performing.\footnote{The differences between \concat and \marqap for both domains are not statistically significant.} Notably, \marqap's average \dlt (\green{$19.6\%$}) is substantially higher compared to the next best method (\green{$14.1\%$}). These results serve as additional strong evidence of \marqap's robustness.
% \\[-15pt]

% ------------------------------------------------------

\paragraph{\marqap's Performance Using Different LMs}
In addition to Longformer, we evaluated \marqap using RoBERTa \cite{ROBERTA}, and BigBird \cite{BigBird} in the standard setting. 
% Since \marqap is prompt-based and independent of the underlying LMreplacing the underlying LM was very portable.
The results are presented in \Cref{tab:model-portability}. \marqap shows a consistent positive effect across different LMs, which further highlights its effectiveness. 

We note that since RoBERTa is a dense-attention Transformer with input length limitation of $512$ tokens, longer passages are split into chunks. This may lead to some chunks containing part of the historic answers, and therefore partial highlighting by \marqap. Our analysis showed that $51\%$ of all examples in QuAC were split into several chunks, and $61\%$ the resulted chunks contained partial highlighting. \marqap's strong performance with RoBERTa suggests that it can remain effective even with partial highlighting.

% ----------------------------------------------------------------------
\begin{table}[t]
\centering
\small
\scalebox{0.7}{
\begin{tabular}{lllc}
\toprule
Model       & No History & \marqapModel & $\Delta$\%  \\ \toprule
RoBERTa & 57.7 & 68.0 & \textbf{(\green{+17.9\%})} \\ 
BigBird   & 57.6  & 66.3 & (\green{+15.1\%}) \\
Longformer$_{base}$   & 60.0 & 68.4 & (\green{+14.0\%}) \\
Longformer$_{squad}$   & \textbf{60.4} & \textbf{70.2} & (\green{+16.6\%}) \\
% Longformer$_{large}$   & \textbf{64.4} & \textbf{72.8} & (+13.0\%) \\
\bottomrule
\end{tabular}
}
\caption{
\marqap's \full setting performance across different Transformer-based pre-trained LMs.
}
% \vspace{-0.2cm}
\label{tab:model-portability}
\end{table}
% % --------------------------------------------------------------------------------

% % ---------------------------------------------------------------------
\begin{table}[t]
\centering
\small
\scalebox{0.7}{
\begin{tabular}{ll}
\toprule
% BiDAF++ \cite{QUAC} & 50.2 \\
BiDAF++ w/ 2-Context \cite{QUAC} & 60.1 \\
HAE \cite{HAE} & 62.4 \\
FlowQA \cite{FlowQA} & 64.1 \\
GraphFlow \cite{GraphFlow} & 64.9 \\
HAM \cite{HAM} & 65.4 \\
FlowDelta \cite{FlowDelta} & 65.5 \\
GHR \cite{GHR} & 73.7 \\
RoR \cite{RoR} & \textbf{74.9} \\
\midrule
\marqapModel (Ours) & 74.0 \\
\bottomrule
\end{tabular}
}
\caption{Results from the official QuAC \daffy leaderboard, presenting F1 scores for the hidden test set, for \marqap and other models with published papers.
}
\vspace{-0.2cm}
\label{tab:leaderboard}
\end{table}
% ----------------------------------------------------------------------

% ------------------------------------------------------
\paragraph{Official QuAC Leaderboard Results}
For completeness, we submitted our best performing model (from the \hr setting) to the official QuAC leaderboard,\footnote{\url{https://quac.ai}.} evaluating its performance on the hidden test set. \Cref{tab:leaderboard} presents the results.\footnote{The leaderboard contains additional results for models which (at the time of writing) include no descriptions or published papers, rendering them unsuitable for fair comparison.}
\marqap achieves a very competitive score of \textbf{74.0} F1, very close to the published state-of-the art (RoR by \citet{RoR} with 74.9 F1), yet with a much simpler model.\footnote{See \S \ref{pre-trained-lm} for a discussion of RoR.}

% ------------------------------------------------------
\subsection{Prompt Design}
\label{sec:ablations}
% \paragraph{Prompt Design}
Recall that \marqap inserts prompts at the beginning and end positions for each historical answer within $P$ (\Cref{fig:method}). The prompts are designed with predefined marker symbols and include the answer's turn index (e.g.\ "<1>").
This design builds on 3 main assumptions: (1) textual prompts can represent conversation history information, (2) the positioning of the prompts within $P$ facilitates highlighting of historical answers,
% serve as answer highlighting mechanism.
and (3) indexing the historical answers encodes valuable information.
We validate our design assumptions by comparing \marqap against ablated variants (\Cref{tab:ablation}).

% % ---------------------------------------------------------------------
\begin{table}[t]
\centering
\small
\scalebox{0.7}{
\begin{tabular}{l|l}
\toprule
% setting & \multicolumn{1}{c}{\Nh} \\
% \midrule

% 100*(58.5-52.9)/52.9
\nohist & 52.9\\
\midrule

\random & 53.8 (\green{+1.7\%}) \\ 
\bare & 59.6 (\green{+12.7\%}) \\ 
\semanticfull & 59.2 (\green{+11.9\%}) \\  
\semantic & 60.4 (\green{+14.2\%}) \\ 
\semantic\emph{+ Index} & 60.7 (\green{+14.8\%}) \\

% \emph{Asymmetric} & 60.9 (\green{+15.1\%}) \\ 
\marqapc  & 61.5 (\green{+16.3\%}) \\ 
\midrule
\marqap & 61.3 (\green{+15.9\%}) \\ 
\bottomrule
\end{tabular}
}
\caption{
% Comparison between \marqap and it's ablated variants, evaluated on the 10\% setting from \Cref{tab:main-table}.
F1 and \dlt scores for \marqap's ablated variants, in the 10\% setup of the \lr setting.
}
\label{tab:ablation}
\vspace{-0.3cm}
\end{table}
% ----------------------------------------------------------------------

To validate assumption (1), we compare \marqap to \marqapc, a variant which adds $\hist_k$ to the input, in addition to \Pmarked and $q_k$. 
\marqapc is exposed to information from $\hist_k$ via two sources: The concatenated $\hist_k$ and the \marqap prompt within \Pmarked.
We observe a negligible effect,\footnote{The difference is not statistically significant.} suggesting that \marqap indeed encodes information from the conversation history, since providing $\hist_k$ does not add useful information on top of \Pmarked.

To validate assumptions (2) and (3), we use two additional \marqap's variants.
\bare inserts a constant predefined symbol ("<>"), in each answer's beginning and end positions within $P$ (i.e. similar to \marqap, but without turn indexing).
\random inserts the same number of symbols but in random positions within $P$.

\bare achieves a \dlt of \green{$12.7\%$}, while \random achieves \green{$1.7\%$}. This demonstrates that the positioning of the prompts within $P$ is crucial, and that most of \marqap's performance gains stem from its prompts positioning w.r.t historical answers \Hanswers. 
When the prompts are inserted at meaningful positions, the model seems to learn to leverage these positions in order to derive an effective history representation.
Surprisingly, \random leads to a minor improvement of \green{$1.7\%$}.\footnote{The difference is statistically significant, we did not further investigate the reasons behind this particular result.} 
Finally, \marqap's improvement over \bare (a \dlt of \green{$15.9\%$} compared to \green{$12.7\%$}), indicates that answer indexing encodes valuable information, helping us validate assumption (3). 

Finally, since textual prompts allow for easy injection of additional information, we make several initial attempts in this direction, injecting different types of information into our textual prompts. In \semantic, the marker contains the first word from the historic answer's corresponding question, which is typically a wh-word (e.g.\ "<what>"). In \semantic\emph{+ Index} we also add the historic answer's turn index (e.g.\ "<what\_1>"). In \semanticfull, we inject the entire historic question into the prompt.
\semantic and \semantic\emph{+ Index} achieved comparable scores, lower than \marqap's but higher than \bare's.\footnote{Both differences are statistically significant.}
This suggests that adding semantic information is useful (since \semantic outperformed \bare), and that combining such information with the positional information is not trivial (since \marqap outperformed \semantic\emph{+ Index}). 
This points at the effects of the prompt structure and the information included, we see that "<1>" and "<what>" both outperform "<>", yet constructing a prompt by naively combining these signals ("<what\_1>") does not lead to complementary effect. 
Finally, \semantic outperformed \semanticfull. We hypothesize that since the full question can be long, it might substantially interfere with the natural structure of the passage text. This provides evidence that the prompts should probably remain compact symbols with small footprint within the passage.
These initial results call for further exploration of optimal prompt design in future work.

% % --------------------------------------------------------------------------------
\subsection{Case Study}
\label{sec:case-study}
\begin{figure}[t]
 \centering
\includegraphics[width=\columnwidth]{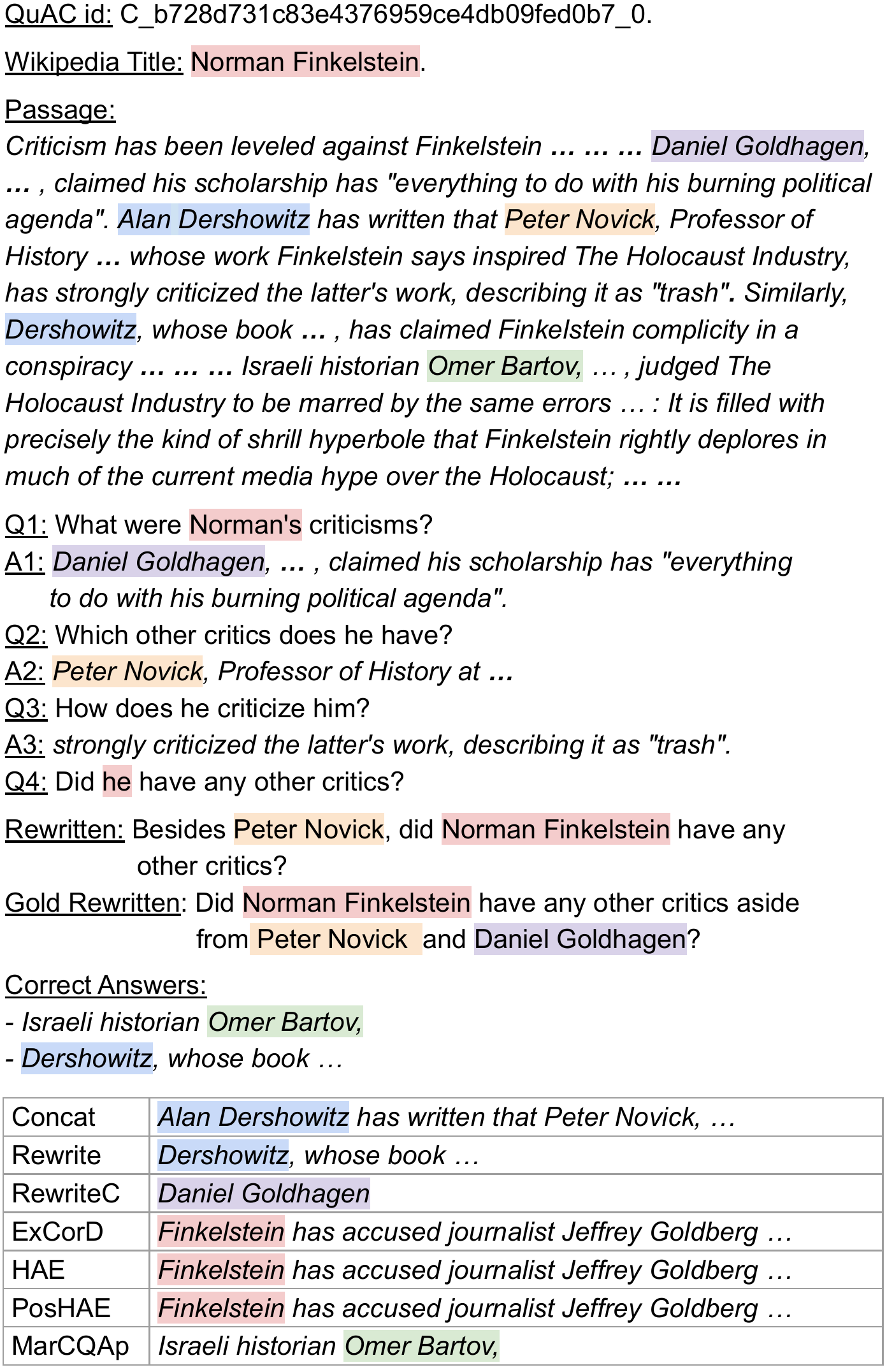}
\vspace{-0.8cm}
\caption{
Our case study example, comparing answers predicted by each evaluated method in the \standard setting. 
We provide a detailed analysis in \S \ref{sec:case-study}.
}
\vspace{-0.4cm}
\label{fig:ex1} 
\end{figure}

\Cref{fig:ex1} presents an example of all evaluated methods in action from the \standard setting. The current question \emph{"Did he have any other \textbf{critics}?"} has two correct answers: \emph{\textbf{Alan Dershowitz}} or \emph{\textbf{Omer Bartov}}.
We first note that all methods predicted a name of a person, which indicates that the main subject of the question was captured correctly. Yet, the methods differ in their prediction of the specific person.

% --------------------------------------------------------------------------------
% --------------------------------------------------------------------------------

\rewrite and \concat predict a correct answer (\emph{Alan Dershowitz}), yet \concat predicts it based on incorrect evidence. This may indicate that \concat did not capture the context correctly (just the fact that it needs to predict a person's name), and was lucky enough to guess the correct name.

Interestingly, \rewritec predicts \emph{Daniel Goldhagen}, which is different from the answers predicted by \concat and \rewrite. This shows that combining both methods can yield completely different results, and demonstrates an instance where \rewritec performs worse than \rewrite and \concat (for instance in the $1\%$ \lr setting).
This is also an example of a history modeling flaw, since \emph{Daniel Goldhagen} was already mentioned as a critic in previous conversation turns.

This example also demonstrates how errors can propagate through a pipeline-based system. The gold rewritten question is \emph{"Did Norman Finkelstein have any other critics aside from \textbf{Peter Novick} and \textbf{Daniel Goldhagen}?"},\footnote{As annotated in CANARD \cite{CANARD}.} while the question rewriting model generated \emph{"Besides \textbf{Peter Novick}, did Norman Finkelstein have any other critics?"}, omitting \textbf{Daniel Goldhagen}. This makes it impossible for \rewrite to figure out that \textbf{Daniel Goldhagen} was already mentioned, making it a legitimate answer. This reveals that \rewrite might have also gotten lucky and provides a possible explanation for the incorrect answer predicted by \rewritec.

\excordlf, \haelf and \poshaelf not only predict a wrong answer, but also seem to fail to resolve the conversational coreferences, since the pronoun \emph{"\textbf{he}"}, in the current question \emph{"Did \textbf{he} have any other critics?"}, refers to \emph{\textbf{Norman Finkelstein}}. 
% which makes their answer unreasonable.

\marqap predicts a correct answer, \emph{Omer Bartov}. This demonstrates an instance where \marqap succeeds while \haelf and \poshaelf fail, even though they are all history-highlighting methods.
Interestingly, \marqap is the only model that predicts \emph{Omer Bartov}, a non-trivial choice compared to \emph{Alan Dershowitz}, since \emph{Omer Bartov} appears later in the passage, further away from the historic answers.

    \section{Limitations}
This work focuses on a single-document CQA setting, which is in line with the majority of the previous work on conversation history modeling in CQA (\S \ref{sec:related}). Correspondingly, \marqap was designed for single-document CQA. Applying \marqap in a multi document settings \cite{OR-QUAC, QReCC, TopiOCQA}
may result in partial history representation, since the retrieved document may contain only part of the historic answers, therefore \marqap will only highlight the answers which appear in the document.\footnote{We note that this limitation applies to all highlighting approaches, including HAE and PosHAE \cite{HAE, HAM}.}

% \marqap represents the conversation history solely by highlighting historic answers in the document, while the retrieved document may contain only part of the historic answers.

In \S \ref{sec:ablations} we showed initial evidence that \marqap prompts can encode additional information that can be useful for CQA. 
% In this work we focused on the core idea behind \marqap as a prompt-based answer highlighting method, 
In this work we focused on the core idea behind prompt-based answer highlighting, 
as a proposed solution in light of our results in \S \ref{sec:study-results}. Yet, we did not conduct a comprehensive exploration in search of the optimal prompt design, and leave this for future work.

\begin{comment}
This work proposes improved and comprehensive evaluation practices for CQA, focusing on robustness to domain shifts and dataset sizes. We chose to follow the commonly adopted metric (word-overlap F1) in CQA, and besides proposing our \delta metric, we did not further explore the idea of improving our metric for better evaluation.
\end{comment}

% \cite{COQA, QUAC, SDNet, FlowQA, FlowDelta, GraphFlow, HAE, HAM, EXCORD, GHR}.
    \section{Conclusion}
\label{sec:conclusion}

In this work, we carry out the first comprehensive robustness study of history modeling approaches for Conversational Question Answering (\DCQA), including sensitivity to model and training data size, domain shift and noisy history input.
We revealed limitations of the existing benchmark-based evaluation, by demonstrating that it cannot reflect the models' robustness to such changes in setting.
% with various methods performing extremely differently under different settings. 
In addition, we proposed evaluation practices that better isolate the contribution of the history modeling component, and demonstrated their usefulness.

We also discovered that highlighting historic answers via passage embedding is very effective in \standard setups, but it
% but it not robust to low resource and domain shift settings.
suffers from substantial performance degradation in low data and domain shift settings.
Following this finding, we design a novel prompt-based history highlighting approach. We show that highlighting with prompts, rather than with embeddings, significantly improve robustness, while maintaining overall high performance.

Our approach can be a good starting point for future work, due to its high effectiveness, robustness and portability. 
% We also hope that our study had shed some light on the limitations of the current benchmark-based evaluations, and that it will encourage more robustness-focused evaluations, leading to better \DCQA systems.
We also hope that the insights from our study will encourage evaluations with focus on robustness,
% more robustness-focused evaluations, l
leading to better \DCQA systems.

\begin{comment}
\cand{
In future work, we'd like to perform similar robustness study to other tasks to expose more fine-grained model behavior. We also plan to explore ways for applying prompt-based highlighting in other tasks, like multi-source summarization, as a compact way of representing relevant information.
}
\end{comment}
    
    % Uncomment for public / camera ready version
    \section*{Acknowledgements}
    We would like to thank the action editor and the reviewers, as well as the members of the IE@Technion NLP group and Roee Aharoni for their valuable feedback and advice. The Technion team was supported by the Zuckerman Fund to the Technion Artificial Intelligence Hub (Tech.AI). This research was also supported in part by a grant from Google.

    % \pagebreak
    \bibliography{anthology,custom}
    \bibliographystyle{acl_natbib}
    
\end{document}